\documentclass{article}

\usepackage[final,main]{neurips_2026}

\usepackage[utf8]{inputenc}
\usepackage[T1]{fontenc}
\usepackage{hyperref}
\usepackage{url}
\usepackage{booktabs}
\usepackage{multirow}
\usepackage{longtable}
\usepackage{amsfonts}
\usepackage{amsthm}
\usepackage{amsmath}
\usepackage{amssymb}
\usepackage{nicefrac}
\usepackage{microtype}
\usepackage[table]{xcolor}
\definecolor{oursrow}{gray}{0.92}
\definecolor{codepink}{HTML}{E6007E}
\usepackage{graphicx}
\usepackage{subcaption}
\usepackage{tikz}
\usetikzlibrary{positioning, arrows.meta, fit, backgrounds, calc}
\usepackage{algorithm}
\usepackage{algorithmic}
\usepackage{soul}

\title{{Two is better than one:} A Collapse-Resistant {Multi-Reward RLIF Training Framework}}

\author{%
  Shourov Joarder$^{1}$ \quad
  Diganta Sikdar$^{1}$ \quad
  Ahsan Habib Akash$^{2}$ \\
  \bf Binod Bhattarai$^{3, 4, 5}$ \quad
  Prashnna Gyawali$^{2}$\thanks{Correspondence to \texttt{prashnna.gyawali@mail.wvu.edu}.} \\[4pt]
  $^{1}$Bangladesh University of Engineering and Technology, Bangladesh \\
  $^{2}$West Virginia University, USA, $^{3}$University of Aberdeen, UK
  \quad \\
  $^{4}$Fogsphere (Redev.AI Ltd, UK),
  $^{5}$ University College London, UK
}

\begin{document}

\maketitle

%% ============================================================
%% ABSTRACT
%% ============================================================
\begin{abstract}
Reinforcement learning with verifiable rewards (RLVR) has substantially improved the reasoning ability of LLMs, but often depends on external supervision from human annotations or gold-standard solutions. Reinforcement learning from internal feedback (RLIF) has recently emerged as a scalable unsupervised alternative, using signals extracted from the model itself. However, existing RLIF methods typically rely on a single internal reward, which can lead to reward hacking, entropy collapse, and degraded reasoning structure. We propose a multi-reward RLIF framework that decomposes the training signal into two complementary components: an answer-level reward based on cluster voting and a completion-level reward based on token-wise self-certainty. To combine these signals robustly, we apply GDPO-based normalization to reduce reward-scale imbalance. We further introduce KL-Cov regularization, which targets the small subset of tokens responsible for disproportionate entropy reduction, preserving exploration and substantially delaying late-stage collapse. Across mathematical reasoning and code-generation benchmarks, our method improves stability and robustness over prior unsupervised RL approaches, while achieving performance close to supervised RLVR methods. These results show that complementary internal rewards, combined with targeted regularization, can support stable long-horizon reasoning without relying on external ground-truth supervision. Code and other details: {\hypersetup{pdfborder={0 0 0}}\textcolor{codepink}{\url{https://machine-intelligence-lab-wvu.github.io/multi-reward/}}}.

\end{abstract}

\begin{figure}[t]
\centering
\begin{subfigure}[b]{0.58\textwidth}
    \includegraphics[width=\linewidth]{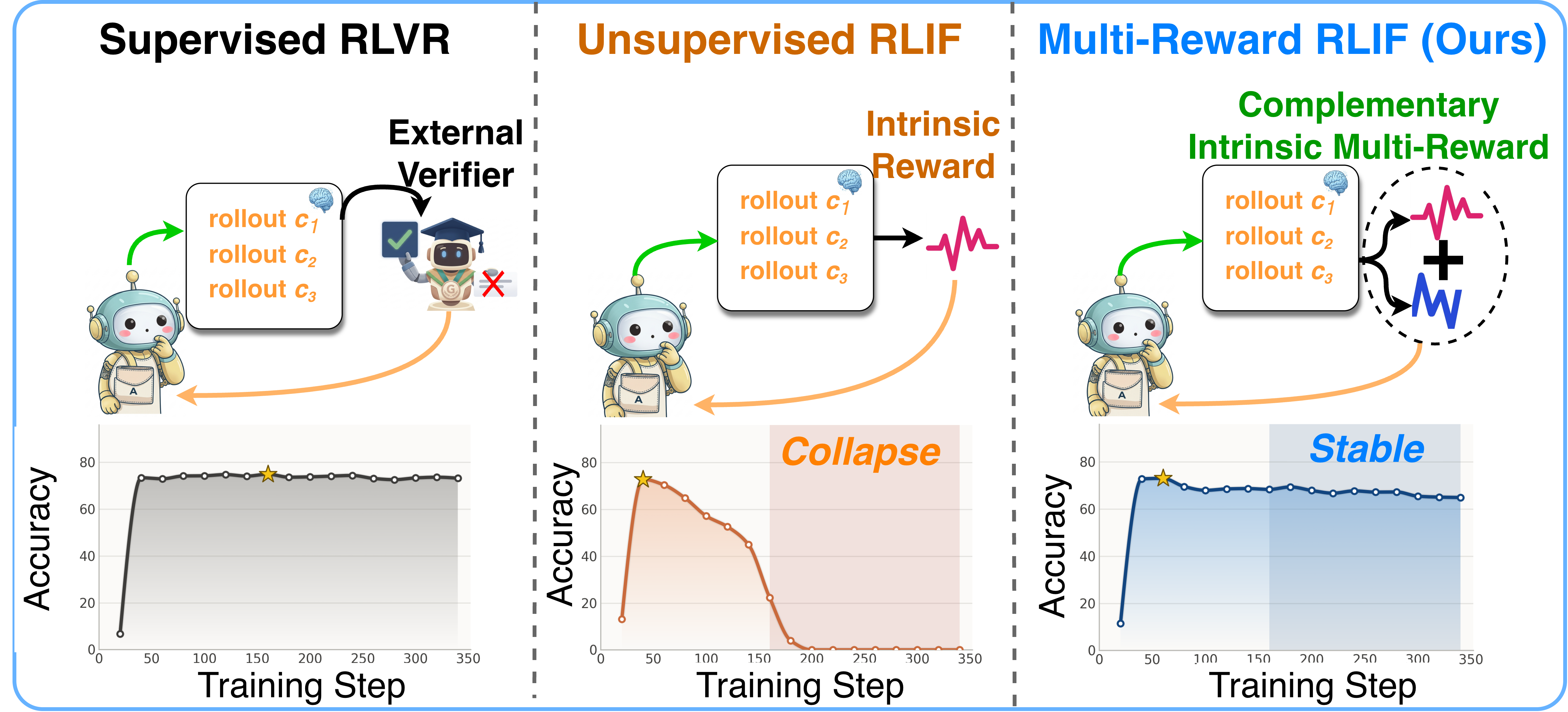}
    \caption{}
    \label{fig:teaser_setting}
\end{subfigure}
\hfill
\begin{subfigure}[b]{0.4\textwidth}
    \includegraphics[width=\linewidth]{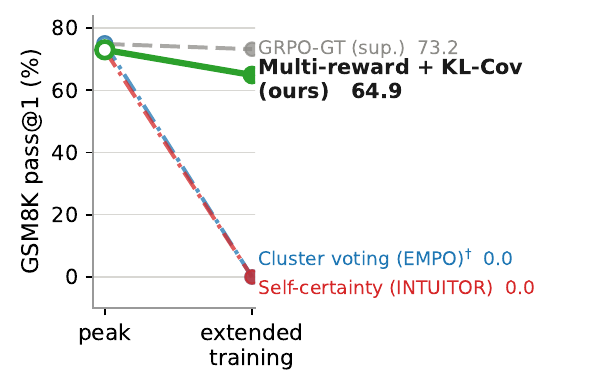}
    \caption{}
    \label{fig:teaser_slope}
\end{subfigure}
\caption{\textbf{RLVR vs.\ single-signal RLIF vs.\ multi-reward RLIF.} (a) Supervised RLVR
requires an external verifier, limiting it to verifiable domains; single-signal RLIF removes
that dependence but collapses under extended training; our multi-reward objective combines
complementary intrinsic signals and stays stable. (b)~GSM8K greedy pass@1 for Qwen2.5-1.5B at each method's best checkpoint and after extended training (step~340; ${}^{\dagger}$~step~280). The single-signal baselines fall to $0.0\%$, whereas multi-reward~+~KL-Cov retains
${\approx}88.9\%$ of its peak, close to supervised GRPO-GT.}
\label{fig:teaser}
\end{figure}

%% ============================================================
%% 1. INTRODUCTION
%% ============================================================
\section{Introduction}
\label{sec:introduction}

Large language models (LLMs) have achieved striking success across domains, yet training them to reason through complex problems remains an open and central challenge in modern AI. Reinforcement learning with verifiable rewards (RLVR) has emerged as a powerful paradigm, enabling models such as DeepSeek-R1~\citep{deepseek_r1} and MiMo~\citep{mimo} to attain strong mathematical reasoning through reward signals derived from ground-truth answer verification. However, RLVR critically depends on external supervision such as human-labeled reasoning traces, verified reference
answers, or pre-trained reward models which limits scalability and practical applicability across domains where such supervision is unavailable, costly, or difficult to obtain.

Reinforcement Learning with Internal Feedback (RLIF) offers an unsupervised alternative that eliminates the need for external supervision entirely.
Instead of relying on ground-truth labels, RLIF methods derive
reward signals solely from intrinsic model signals: either through consensus-based measures such as majority voting and semantic clustering~\citep{empo}, or through entropy minimization measures such as the self-certainty score~\citep{intuitor}.
These approaches have demonstrated that meaningful reasoning improvements are achievable without 
labeled data, making them 
a promising path toward 
scaling LLM reasoning to arbitrary domains~\citep{ttrl, absolute_zero, maximizing_confidence}.

Despite these promising results, existing RLIF methods exhibit recurring failure patterns, including reward hacking and model collapse~\citep{entropy_effectiveness, maximizing_confidence}. Prior work has observed that models often achieve strong initial gains but gradually degrade after extended training; Figure~\ref{fig:teaser} contrasts the three settings (RLVR, naive RLIF, and our multi-reward RLIF) and shows the resulting gap at the end of the horizon. We argue that this decline stems from the structure of the reward signals themselves. {For example, self-certainty rewards high-confidence token distributions, effectively using entropy minimization as a proxy for output quality. Majority voting, though consensus-based rather than explicitly entropy-driven, similarly favors outputs that the model already assigns high probability to, thereby indirectly reinforcing low-entropy behavior. Under both signals, the policy is incentivized to reduce token-level uncertainty, whether or not this reduction improves reasoning quality.}

This incentive creates two compounding failure modes. First, high-entropy transitional tokens, such as \emph{however}, \emph{thus}, \emph{because}, and \emph{suppose}, which often serve as structural anchors in chain-of-thought reasoning, are disproportionately penalized~\citep{nfl}. This can weaken logical coherence even when final-answer accuracy temporarily improves. Second, because shorter responses contain fewer uncertain tokens, the reward implicitly favors brevity over depth~\citep{he2026unsupervised}, encouraging terse, high-probability outputs rather than genuine improvements in reasoning quality. This pattern of early improvement followed by degradation is an empirical instance of Goodhart’s Law~\citep{gao2022scalinglawsrewardmodel}: when a single proxy becomes the sole optimization target, the policy learns to exploit the proxy rather than improve the underlying capability. As optimization proceeds, the reward signal and true task performance can decouple, leading to overoptimization.

This motivates a multi-reward RLIF objective that jointly optimizes structurally complementary signals rather than relying on a single proxy. Specifically, we propose a multi-reward RLIF framework that combines answer agreement and model confidence as intrinsic rewards.
To prevent either reward from dominating optimization, we normalize the advantages induced by each reward independently.
We further introduce a targeted regularization strategy to reduce collapse during optimization. In particular, to mitigate entropy collapse while preserving exploration, we use KL-Cov regularization, inspired by~\citet{cui2025entropy}, which targets the small subset of tokens responsible for disproportionate entropy reduction while maintaining diversity in the remaining token distribution.
Across five diverse mathematical reasoning benchmarks and two coding benchmarks, we show that our method outperforms several baselines, including single-reward RLIF frameworks such as self-certainty and majority voting \citep{intuitor, ttrl} while maintaining stable gains under continued training. This stability is crucial for RLIF: as models scale and reasoning trajectories grow more complex, training must preserve exploration beyond early high-confidence behaviors rather than collapse into narrow reward-seeking patterns.

Our contributions can be summarized as follows:
\begin{itemize}
    
    \item We propose a multi-reward RLIF framework with GDPO normalization that jointly optimizes two complementary signals: answer-level consensus and completion-level self-certainty, mitigating the failure modes of single-proxy optimization.

    \item We adapt KL-Cov regularization to the unsupervised multi-reward setting as a targeted policy constraint on high-covariance, entropy-collapsing tokens, preserving exploration over the remaining token distribution.
  
    \item Across five mathematical reasoning benchmarks and two coding benchmarks, we show that our method improves performance over single-reward RLIF baselines and remains stable under continued training.
    
\end{itemize}

%% ============================================================
%% 2. RELATED WORK
%% ============================================================
\section{Related Work}
\label{sec:related}
\textbf{Reinforcement learning with internal feedback (RLIF):}
To eliminate external supervision for RL training of LLMs, recent work has explored RLIF, where the reward is derived from the model's intrinsic signals.
Self-Certainty (INTUITOR)~\citep{intuitor} rewards more confident reasoning traces using the average KL divergence between a uniform distribution and the model's next-token distribution, while EMPO~\citep{empo} takes a complementary approach, semantically clustering rollout answers and rewarding them probabilistically based on their cluster size.
TTRL~\citep{ttrl} shows that majority voting yields surprisingly effective rewards for test-time RL without ground-truth labels.
Absolute Zero~\citep{absolute_zero} and R-Zero~\citep{rzero} propose self-play frameworks where models generate their own training tasks; \citet{verifree} extend RL-based reasoning to general domains without verifiers by maximizing the probability of generating reference answers; Co-rewarding~\citep{corewarding} seeks complementary supervision from multiple views to stabilize self-supervised RL; and SSR-Zero~\citep{ssr_zero} extends self-rewarding RL to machine translation, demonstrating generalization beyond mathematical reasoning.
While these methods advance unsupervised RL in various directions, most of them rely on a single reward signal or feedback mechanism, making them susceptible to the failure modes described in Section~\ref{sec:introduction}.

\textbf{Entropy minimization, entropy collapse, and model collapse:}
Uncontrolled entropy minimization, rewarding more confident outputs to improve reasoning without any labels~\citep{entropy_effectiveness, maximizing_confidence}, is also the primary driver of collapse. \citet{exploration_exploitation} formalize how clipping bias under spurious rewards reduces policy entropy, and \citet{cui2025entropy} show that ${\sim}94\%$ of entropy reduction happens in the earliest training steps, driven by a tiny ${\sim}0.02\%$ fraction of high-covariance tokens (anomalously large centered cross-products between log-probability and advantage) that dominate the policy gradient and drive entropy to zero. The downstream pathology is well-documented: \citet{srt} report complete model collapse in self-training (identical outputs regardless of input), \citet{nfl} show that RLIF approaches exhibit a rise-then-fall trajectory from over-optimizing a single internal signal, and \citet{he2026unsupervised} confirm the pattern at scale, with collapse timing set by model priors.

\textbf{Multi-objective RL and reward combination:}
Combining multiple rewards is challenging because naive summation leads to scale dominance: the larger-variance reward overwhelms the gradient.
GDPO~\citep{gdpo} addresses this by normalizing each reward channel independently within groups before combination.
Our work applies this principle to combining answer-level and completion-level rewards for unsupervised LLM training, paired with covariance-aware regularization to mitigate entropy collapse.

%% ============================================================
%% 4. METHOD
%% ============================================================
\begin{figure}[t]
    \centering
    \includegraphics[width=\textwidth]{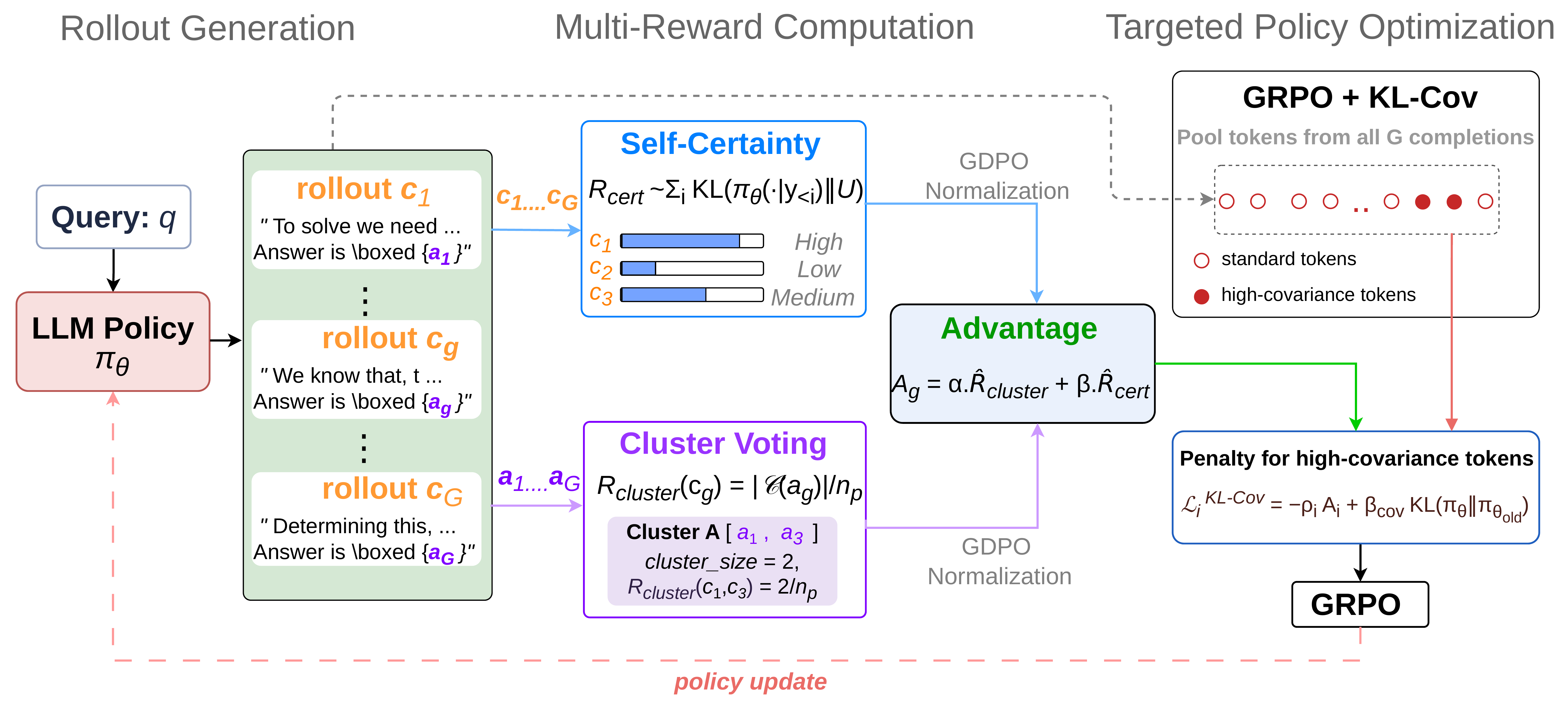}
    \caption{\textbf{Multi-reward + KL-Cov framework}: Given a prompt, we sample G completions from the rollout policy and compute two intrinsic rewards: an answer-level cluster-voting reward over final-answer agreement, and a completion-level self-certainty reward over token-level predictive distributions. Each reward is normalized within the prompt group before being combined into rollout-level advantages. Optimization uses a GRPO-style objective, while KL-Cov penalizes only high-covariance tokens to regularize concentrated updates.}
    \label{fig:framework}
\end{figure}

\section{Method}
\label{sec:method}

We present multi-reward + KL-Cov, a multi-reward RLIF training framework that combines two complementary reward signals: answer-level agreement and completion-level self-certainty with group-wise reward normalization and covariance-aware regularization, within a GRPO-style policy optimization objective.

{At a high level, for each prompt $q$, our method samples $G$ completions from the rollout policy $\pi_{\theta_{\mathrm{old}}}$, extracts each final answer, and clusters the answers by consensus. Each completion receives two rewards: an answer-agreement reward determined by the size of its answer cluster, and a completion-level confidence reward computed using self-certainty. We normalize each reward channel within the $G$-sample group, form a rollout-level advantage from their weighted sum, and optimize the policy with a GRPO-style objective, with additional log-ratio regularization applied to high-covariance tokens.} Figure~\ref{fig:framework} illustrates the overall training pipeline.

\subsection{Preliminaries}
\label{sec:preliminaries}

Let $\pi_\theta$ denote the current policy, $\pi_{\theta_{\mathrm{old}}}$ denote the rollout policy used to generate the current training batch, and $\pi_{\mathrm{ref}}$ denote a frozen reference policy. Given a prompt $q$, we sample $G$ completions from $\pi_{\theta_{\mathrm{old}}}$. The $g$-th completion is denoted by $c_g = (c_{g,1}, \ldots, c_{g,T_g})$, where $T_g$ is its length, and its extracted final answer is denoted by $a_g$. During optimization, each token in completion $c_g$ inherits the rollout-level advantage $A_g$ computed from the reward signals.

Flattening across rollouts, we write $\{y_1, \ldots, y_N\}$ for the $N$ tokens in a training batch, where $y_i$ inherits the advantage $A_i$ of the rollout containing it.
We optimize the policy using a GRPO-style clipped surrogate objective~\citep{grpo}. For a generated token $y_i$ with inherited advantage $A_i$, let
$\rho_i = \pi_\theta(y_i \mid q, y_{<i}) / \pi_{\theta_{\mathrm{old}}}(y_i \mid q, y_{<i})$
be the importance sampling ratio between the current and rollout policies. The per-token GRPO loss is
\begin{equation}
\label{eq:grpo_standard}
\mathcal{L}_i^{\text{GRPO}}(\theta) = -\min\!\Big[\rho_i A_i,\; \text{clip}(\rho_i,\, 1{-}\epsilon,\, 1{+}\epsilon)\, A_i\Big]
\end{equation}
where $\epsilon > 0$ is the clipping parameter.

\subsection{Answer-Level Reward via Cluster Voting}
\label{sec:cluster_voting}

The first reward channel assigns higher reward to completions whose final answers agree with many other sampled completions for the prompt: given $G$ rollouts for a prompt $q$, we extract the final answers $a_1, \ldots, a_G$, group them into clusters by equivalence on the normalized answer string, and, rather than binary majority voting, which rewards only the most popular answer, assign each rollout $g$ a reward proportional to the fraction with equivalent answers:

\begin{equation}
    \label{eq:cluster_reward}
    R_{\text{cluster}}(a_g) =
    \begin{cases}
    \big|\{g' \in \{1,\ldots,G\} : a_{g'} \equiv a_g\}\big|\,/\,n_{\text{p}} & \text{if } a_g \text{ is
  parseable},\\
    0 & \text{otherwise,}
    \end{cases}
  \end{equation}
 
where $n_{\text{p}}$ is the number of rollouts (out of $G$) whose answer string parses successfully. For mathematical problems, the equivalence is measured with numerical consistency.
This provides a denser signal than binary majority voting: minority clusters receive small but nonzero rewards rather than zero; such clustering is robust to surface-level variation in equivalent answers~\citep{empo, semantic_uncertainty}.

In implementation, we additionally include a small auxiliary boxed-format reward (weight $0.1$) alongside the cluster reward to gate non-parseable completions; the two are summed and normalized together as one channel (reported in Appendix~\ref{app:hyperparams}).

\subsection{Completion-Level Reward via Self-Certainty}
\label{sec:self_certainty}
Following \citet{intuitor}, we use self-certainty as a completion-level reward. Let $\mathcal{V}$ denote the vocabulary and let $\mathcal{U}$ denote the uniform distribution over $\mathcal{V}$. For a completion $c_g = (c_{g,1}, \ldots, c_{g,T_g})$, we compute the self-certainty reward online from the current policy as
\begin{equation}
\label{eq:cert_reward}
R_{\text{cert}}(c_g) = \frac{1}{T_g} \sum_{t=1}^{T_g} D_{\text{KL}}\!\Big(\mathcal{U} \;\Big\|\; \pi_\theta(\cdot \mid q, c_{g,<t})\Big)
\end{equation}
where $c_{g,<t} = (c_{g,1}, \ldots, c_{g,t-1})$ is the prefix before token $t$ of generation $g$.
{Used in isolation, self-certainty 
% may 
has been shown to 
favor degenerate policies that produce overly deterministic or short templated completions regardless of the input~\citep{srt, nfl, he2026unsupervised}. 
% (also in Appendix \ref{app:qualitative}). 
Our qualitative examples (Appendix~\ref{app:qualitative}) show the same reward-hacking, and our single-reward decomposition (Figure~\ref{fig:single_multi_reward}) confirms it: self-certainty alone collapses to ${\sim}100$-token templates with token-distribution entropy in $[0.03, 0.17]$.} 

This motivates our \textit{multi-reward} design: rather than optimizing either signal independently, multi-reward + KL-Cov jointly optimizes two complementary signals---self-certainty, which captures the policy's confidence in generated tokens, and cluster voting, which captures the agreement among sampled final answers.
 
\subsection{Group-Wise Multi-Reward Normalization}
\label{sec:gdpo}

Since the two rewards have different numerical scales (cluster-voting lies in $[0, 1]$ while self-certainty depends on vocabulary size), naive combination can let one signal dominate the advantage purely through scale, degrading performance. We confirm this empirically in Appendix~\ref{app:gdpo_ablation}: \textit{combine-then-normalize} degrades downstream accuracy (Figure~\ref{fig:gdpo_ablation}), and a logprob-space gradient diagnostic (Figure~\ref{fig:gdpo_gradients}) shows the self-certainty channel dominating the per-token gradient by roughly an order of magnitude while the cluster channel is silenced.
To avoid this, we normalize each reward channel independently within the group of $G$ completions for the same prompt, following the GDPO principle~\citep{gdpo}. 
For each reward channel $k \in \{\text{cluster}, \text{cert}\}$, $R_k(g)$ is the
reward of rollout $g$, and we compute the group mean and standard deviation as $\mu_k = \frac{1}{G}\sum_{g=1}^{G} R_k(g)$ and $\sigma_k = \text{std}(\{R_k(g)\}_{g=1}^{G})$.
The normalized reward for rollout $g$ under channel $k$ is:
\begin{equation}
\label{eq:gdpo_norm}
\hat{R}_k(g) = \frac{R_k(g) - \mu_k}{\sigma_k + \delta}
\end{equation}
where $\delta > 0$ is a small constant for numerical stability.

We then combine the normalized rewards into a single rollout-level advantage:
\begin{equation}
\label{eq:advantage}
A_g = \alpha \cdot \hat{R}_{\text{cluster}}(g) + \beta \cdot \hat{R}_{\text{cert}}(g)
\end{equation}
where $\alpha, \beta$ {$\geq$} $0$ are reward weights. The weights $\alpha$ and $\beta$ control the relative importance of answer quality versus reasoning quality; {for this study we empirically set $\alpha = \beta = 0.5$}.

\subsection{KL-Cov: Covariance-Aware Policy Optimization}
\label{sec:klcov}

The two rewards provide complementary supervision, but each induces a different degeneration mode when optimized alone: cluster-only training drifts in answer space (rising entropy, lengthening completions), while training with self-certainty collapses token-space entropy into short ${\sim}100$-token templates (Figure~\ref{fig:single_multi_reward}). Although GDPO normalization mitigates these instabilities, the resulting policy degenerates around step~240 when trained without additional exploration regularization (Figure~\ref{fig:klcov_ablation}).

This motivates our covariance-aware targeted regularization. We adopt KL-Cov, a regularizer that targets the tokens responsible for collapse without unnecessarily constraining the rest of the policy: {in each batch, the top-$k$ ($k{=}0.02$) tokens by covariance receive a targeted KL penalty, leaving the remaining ${\sim}98\%$ under the standard clipped GRPO surrogate.}
Using the set of $N$ batch tokens $\{y_1, \ldots, y_N\}$ defined in Section~\ref{sec:preliminaries}, we compute 
{a per-token covariance contribution between token log-probabilities and rollout advantages as the centered cross-product of the two:

\begin{equation}
\label{eq:covariance}
\text{Cov}(y_i) = \left(\log \pi_\theta(y_i \mid y_{<i}) - \frac{1}{N}\sum_{j=1}^{N} \log \pi_\theta(y_j \mid y_{<j})\right) \cdot \left(A_i - \frac{1}{N}\sum_{j=1}^{N} A_j\right)
\end{equation}
{A large positive $\text{Cov}(y_i)$ means the token is either more probable than average with higher-than-average advantage (encouraging entropy collapse), or less probable than average with lower-than-average advantage (contributing no effective gradient learning).} We rank all $N$ tokens by covariance and select the top-$k$ fraction as the high-covariance set:
\begin{equation}
\label{eq:klset}
\mathcal{I}_{\text{KL}} = \big\{i \in \{1, \ldots, N\} \;\big|\; \text{Rank}\!\big(\text{Cov}(y_i)\big) \leq k \cdot N\big\}, \quad k \ll 1
\end{equation}
where $\text{Rank}(\cdot)$ orders tokens from highest to lowest covariance, so rank 1 is the most problematic token. The per-token loss depends on whether the token belongs to $\mathcal{I}_{\text{KL}}$:
\begin{equation}
\label{eq:per_token_loss}
\mathcal{L}(y_i; \theta) = \begin{cases}
-\min\!\Big[\rho_i A_i,\; \text{clip}(\rho_i, 1{-}\epsilon, 1{+}\epsilon)\, A_i\Big] & \text{if } i \notin \mathcal{I}_{\text{KL}} \\[6pt]
-\rho_i A_i + \beta_{\text{cov}} \big|\log \pi_\theta(y_i \mid y_{<i}) - \log \pi_{\theta_{\text{old}}}(y_i \mid y_{<i})\big| & \text{if } i \in \mathcal{I}_{\text{KL}}
\end{cases}
\end{equation}
The first case is the standard clipped GRPO surrogate (Eq.~\ref{eq:grpo_standard}).
The second case replaces clipping with a targeted penalty: the term $|\log \pi_\theta(y_i \mid y_{<i}) - \log \pi_{\theta_{\text{old}}}(y_i \mid y_{<i})| = |\log \rho_i|$ is the absolute log-importance-ratio, an $\ell_1$ surrogate for the token-level KL between $\pi_{\theta_{\text{old}}}$ and $\pi_\theta$ that mirrors the targeted KL penalty of \citet{cui2025entropy} while remaining symmetric in the direction of the shift; $\beta_{\text{cov}} > 0$ controls the penalty strength.

\noindent
\textbf{Total Loss:} The total batch loss, with a global KL penalty against the frozen reference policy $\pi_{\text{ref}}$, is:
\begin{equation}
\label{eq:total_loss}
\mathcal{L}(\theta) = \frac{1}{N}\sum_{i=1}^{N} \mathcal{L}(y_i; \theta) \;+\; \gamma\, D_{\text{KL}}(\pi_\theta \| \pi_{\text{ref}})
\end{equation}
where $\gamma \geq 0$ is the reference KL coefficient of GRPO.
{Following {\citet{cui2025entropy}}, we use $k = 0.02$ ($2\%$); the targeted penalty mitigates entropy collapse while preserving exploration for the ${\sim}98\%$ of tokens under the standard update.}

%% ============================================================
%% 5. EXPERIMENTS
%% ============================================================

\section{Experiments \& Results}
\label{sec:experiments}

\subsection{Experimental Setup}
\label{sec:setup}
\begin{figure*}[t]
\centering
\includegraphics[width=1\textwidth]{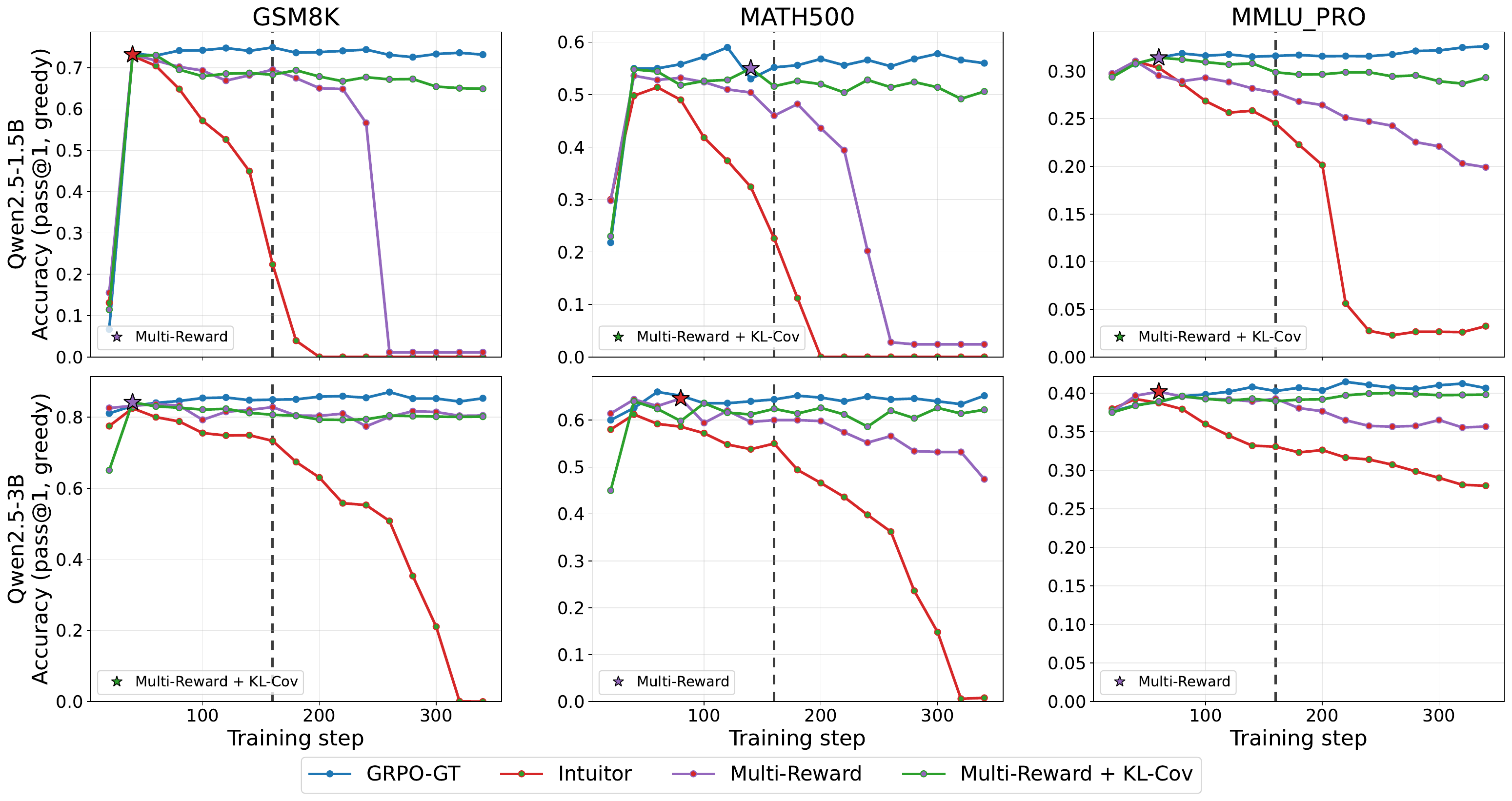}
\caption{\textbf{Our method overcomes model collapse in extended training following supervised GRPO-
GT trajectory}. Accuracy (pass@1) trajectories on GSM8K, MATH500, MMLU-Pro for Qwen2.5-1.5B (top) and Qwen2.5-3B (bottom). The dashed vertical line marks step~160 (Table~\ref{tab:main_results_all}); the star ($\bigstar$) marks the peak among unsupervised methods (GRPO-GT excluded), labeled in-axis. Only multi-reward~+~KL-Cov tracks GRPO-GT through the full 340-step horizon at both scales.}
\label{fig:trajectory}
\end{figure*}

\paragraph{Training details.}

We train Qwen2.5-Base (1.5B, 3B, and 7B)~\citep{qwen2.5}, and Qwen3-1.7B-Base for all experiments. For training data, we use the MATH training split of \texttt{MATH-lighteval} ~\citep{hendrycksmath2021}, as the source of unlabeled training prompts. All models are trained with full finetuning using GRPO~\citep{grpo} as the base optimizer with a learning rate of $3 \times 10^{-6}$ and cosine scheduling.
We use $G = 7$ rollouts per prompt with a maximum completion length of 3072 tokens. The GRPO KL penalty coefficient is set to $\gamma = 0.005$, following \citep{intuitor}.
For our method, we set equal reward-channel weights $\alpha=\beta=0.5$, use a top-$2\%$ covariance threshold $k=0.02$, and set $\beta_{\mathrm{cov}}=0.05$. Training is conducted on a single node with $4$ NVIDIA GH200 96GB GPUs. The effective batch size is $882$ completions ($126$ prompts $\times$ $7$ rollouts) per gradient update. We train for $340$ steps using \texttt{huggingface-trl} framework \citep{vonwerra2020trl}. The full training, optimization, batch, hardware, and evaluation hyperparameters are reported in Appendix~\ref{app:hyperparams} and Table~\ref{tab:hyperparams}.

\paragraph{Baselines.}
{Our baselines cover the two dominant RLIF signal families. \emph{INTUITOR}~\citep{intuitor} is the canonical entropy-minimization method. Our \emph{cluster-only} implementation represents the consensus/majority-voting family: on math it optimizes the same objective as EMPO~\citep{empo}, whose implementation also falls back to exact-match clustering for numeric answers. TTRL~\citep{ttrl} uses the same vote signal but trains on the test stream, which is a different protocol; LLM-as-judge self-rewarding employs an explicit generative reward model and is out of scope. \emph{GRPO-GT}~\citep{grpo} is the supervised upper bound.}

\paragraph{Evaluation.}
We perform extensive evaluation on seven benchmarks spanning from in-distribution to out-of-distribution (OOD) reasoning: GSM8K~\citep{cobbe2021gsm8k}, MATH500~\citep{hendrycksmath2021}, MMLU-Pro~\citep{wang2024mmlupro} (5-shot CoT), AIME 2024/2025, 
% (30 problems each; included as far-OOD probes), 
LiveCodeBench v6~\citep{livecodebench} (LCB; pass@1, all-tests-must-pass), and CRUXEval-O~\citep{cruxeval} (output-prediction via ZeroEval~\citep{zeroeval} \texttt{OEQA} template). 
Math benchmarks are scored with \texttt{math\_verify} using the training chat-style prompt; MMLU-Pro uses TIGER-Lab's canonical 5-shot CoT completion-style prompt~\citep{wang2024mmlupro}.
LCB evaluation follows the lighteval~\citep{lighteval} pipeline verbatim. Among these, AIME 2024/2025 and the two coding tasks serve as OOD generalization, as olympiad-level reasoning and code execution are not present in training dataset.

\subsection{Main Results}
\label{sec:main_results}
\paragraph{Math-reasoning Performance.}
Table~\ref{tab:main_results_all} reports greedy ($T{=}0$, $n{=}1$) pass@1 at a fixed step-160 checkpoint across three math benchmarks for both Qwen2.5-1.5B and Qwen2.5-3B.{
We select step 160 as a representative mid-training checkpoint,
% (18 epochs), 
prior to the late-stage collapse dynamics observed in single-signal continued training; full trajectories are provided in Figure \ref{fig:trajectory}.
}
On every math benchmark, the best unsupervised method is one of our two multi-reward variants. On 1.5B, multi-reward~+~KL-Cov leads INTUITOR by $45.9$, $29.0$, and $5.4$ pp on GSM8K, MATH500, and MMLU-Pro, and trails supervised GRPO-GT by only $6.6$, $3.6$, and $1.7$ pp. The gains are consistent across model scales: on the 3B model both variants also outperform INTUITOR and remain on par with GRPO-GT. {The results on AIME 2024/2025 are reported in Appendix~\ref{app:aime}, and a third-scale trajectory study on Qwen2.5-7B is reported in Appendix~\ref{app:7b}.}

\begin{table}[t]
\caption{Pass@1 (\%, greedy $T{=}0$, $n{=}1$) on three math-reasoning and two code-reasoning benchmarks at the step-160 checkpoint. Best \emph{unsupervised} value per column in \textbf{bold}.}
\label{tab:main_results_all}
\centering
\small
\resizebox{\textwidth}{!}{%
\begin{tabular}{llccccc}
\toprule
& & \multicolumn{3}{c}{\textbf{Math-reasoning}} & \multicolumn{2}{c}{\textbf{Code-reasoning}} \\   
  \cmidrule(lr){3-5} \cmidrule(lr){6-7}
\textbf{Model} & \textbf{Method} & \textbf{GSM8K} & \textbf{MATH500} & \textbf{MMLU-Pro} & \textbf{LCB v6} & \textbf{CRUXEval-O} \\
\midrule
\multirow{4}{*}{Qwen2.5-1.5B}
 & GRPO with ground-truth reward                 & 74.9 & 55.2 & 31.6 & 1.7 & 24.9\\
 & INTUITOR                         & 22.4 & 22.6 & 24.5 & 0.8 & 12.6\\
 & \cellcolor{oursrow}Multi-reward (ours) & \cellcolor{oursrow}\textbf{69.5} & \cellcolor{oursrow}46.0 & \cellcolor{oursrow}27.7 & \cellcolor{oursrow}\textbf{3.3} & \cellcolor{oursrow}\textbf{21.8} \\
 & \cellcolor{oursrow}Multi-reward + KL-Cov (ours) & \cellcolor{oursrow}68.3 & \cellcolor{oursrow}\textbf{51.6} & \cellcolor{oursrow}\textbf{29.9} & \cellcolor{oursrow}2.6 & \cellcolor{oursrow}20.5 \\
\midrule
\multirow{4}{*}{Qwen2.5-3B}
 & GRPO with ground-truth reward                 & 84.9 & 64.4 & 40.3 & 6.4 & 41.0\\
 & INTUITOR                          & 73.3 & 55.0 & 33.1 & 6.6 & 36.1\\
 & \cellcolor{oursrow}Multi-reward (ours) & \cellcolor{oursrow}\textbf{82.8} & \cellcolor{oursrow}60.0 & \cellcolor{oursrow}\textbf{39.3} & \cellcolor{oursrow}4.4 & \cellcolor{oursrow}\textbf{41.0} \\
 & \cellcolor{oursrow}Multi-reward + KL-Cov (ours) & \cellcolor{oursrow}80.7 & \cellcolor{oursrow}\textbf{62.4} & \cellcolor{oursrow}39.0 & \cellcolor{oursrow}\textbf{7.7} & \cellcolor{oursrow}38.3 \\
\bottomrule
\end{tabular}%
}
\end{table}

\paragraph{Code-reasoning generalization.}
We evaluate code-reasoning generalization to assess whether math-trained reasoning structure transfers to code-execution domains without any code-domain training. Table~\ref{tab:main_results_all} reports results on two coding benchmarks, LCB v6 and CRUXEval-O, evaluated at the same step-160 checkpoints. On Qwen2.5-3B, multi-reward (with or without KL-Cov) tracks supervised GRPO-GT closely: plain multi-reward matches GRPO-GT exactly on CRUXEval-O ($41.0\%$), and multi-reward + KL-Cov is the strongest unsupervised method on LCB ($7.7\%$, ahead of GRPO-GT at $6.4\%$ and INTUITOR at $6.6\%$). On Qwen2.5-1.5B all methods sit near floor on LCB ($0.8$--$3.3\%$) so the pattern is dominated by base-model capability; on CRUXEval-O the same INTUITOR-vs-multi-reward separation as the math benchmarks holds ($12.6\%$ vs.\ $21.8$/$20.5\%$).

\paragraph{Complementary intrinsic reward signals resists model collapse:}
\label{sec:entropy_analysis}

In Figure~\ref{fig:trajectory}, we illustrate model collapse by showing the full training dynamics, with checkpoints evaluated every 20 training steps. We report results on GSM8K (in-distribution arithmetic), MATH500 (in-distribution competition math), and MMLU-Pro (OOD multiple-choice reasoning), for both Qwen2.5-1.5B and Qwen2.5-3B.
Across these settings, we observe two consistent patterns at both scales. First, single-reward INTUITOR exhibits model collapse due to reward-hacking (See Appendix \ref{app:qualitative}): self-certainty peaks early, around steps~40--60, and then declines monotonically, reaching near-zero math accuracy. Second, ours \textbf{Multi-reward~+~KL-Cov} remains stable across the full training trajectory. It performs on par with supervised GRPO-GT, retaining $89$--$93\%$ of peak performance on 1.5B and $95$--$99\%$ on 3B at step~340. We also observe that multi-reward training alone delays collapse but does not fully prevent it, especially on 1.5B. {The cluster-only training, which on math is the EMPO objective (Section~\ref{sec:setup}), shows complementary collapse by exploiting the answer-level reward, after which entropy and response length rise (Figure~\ref{fig:single_multi_reward}).}

{\paragraph{Collapse resistance across scales and priors.}
To test whether these conclusions are an artifact of the 1.5B/3B setting, we repeat the trajectory study on Qwen2.5-7B (scale) and Qwen3-1.7B-Base (newer pretraining generation) with the same recipe. The single-signal failure \emph{intensifies} with capability: on 7B, INTUITOR collapses totally by step~120 ($0.0\%$ even on pass@$K$), while ours shows no decline phase and matches GRPO-GT under sampling; under the Qwen3 prior the failure appears as sustained decay, with INTUITOR dropping below the base model on MATH500 while ours retains $97.8$/$95.8\%$ of peak and is statistically indistinguishable from GRPO-GT on MATH500/MMLU-Pro (Appendices~\ref{app:7b}--\ref{app:qwen3}). Table~\ref{tab:scaling} summarises retention across all four backbones.}

\begin{table}[t]
\caption{{\textbf{Collapse resistance holds across scales and priors.} Retention $=$ end\,/\,peak pass@1 (\%) on GSM8K\,/\,MATH500; horizon in parentheses. \textbf{Bold}: best unsupervised. Sources: Table~\ref{tab:stability}, Appendices~\ref{app:7b}--\ref{app:qwen3}.}}
\label{tab:scaling}
\centering
\small
\begin{tabular}{lccc}
\toprule
\textbf{Backbone (horizon)} & \textbf{GRPO-GT (sup.)} & \textbf{INTUITOR} & \textbf{Multi-Reward + KL-Cov (Ours)} \\
\midrule
Qwen2.5-1.5B (340)    & 97.7 / 94.9 & 0.0 / 0.0   & \textbf{88.9 / 92.0} \\
Qwen2.5-3B (340)      & 98.0 / 98.8 & 0.0 / 1.3   & \textbf{95.2 / 97.2} \\
Qwen2.5-7B (240)      & 99.1 / 96.6 & 0.0 / 0.0   & \textbf{100.0 / 97.6} \\
Qwen3-1.7B-Base (200) & 99.0 / 97.9 & 80.9 / 81.5 & \textbf{97.8 / 95.8} \\
\bottomrule
\end{tabular}
\end{table}

\subsection{Ablation Study}
\label{sec:ablation}

We isolate the contribution of each component with two trajectory-level ablations: a single-reward decomposition that targets the complementary-rewards claim, and a $\beta_{\text{cov}}$ sweep that targets sensitivity to the KL-Cov coefficient.  
The component-removal table, the INTUITOR$+$KL-Cov trajectory, and the full $\beta_{\text{cov}}$ tabulation are deferred to Appendix~\ref{app:component_ablation} and~\ref{app:klcov_sweep}.

\paragraph{Single- \& Multi-reward 
analysis without KL-Cov.}
% decomposition: complementary failure modes.}
We analyze the effect of single- and multi-reward structures in the absence of KL-Cov regularization by comparing three reward settings based on cluster-voting, self-certainty, and their combination. 
All three settings are initialized from Qwen2.5-1.5B-Base and trained for 280 steps. Figure~\ref{fig:single_multi_reward} reports pass@1 accuracy together with the corresponding mean completion length and entropy.

\emph{Cluster-only} training attains the highest early GSM8K accuracy ($75.0\%$ at step~60), but collapses abruptly, with GSM8K dropping from $71.8\%$ to $0.0\%$ between steps~120 and~140 and MATH500 from $35.4\%$ to $1.4\%$. Since entropy and completion length increase after collapse, the model produces long completions whose extracted answers collapse to a dominant cluster, exploiting the answer-level reward. In contrast, \emph{self-certainty-only} (\textit{i.e.,} INTUITOR) training induces token-mode (or entropy) collapse: entropy falls from $2.4$ to $0.08$ {nats} by step~60, and completion length shrinks from ${\sim}990$ to ${\sim}100$ tokens. The vulnerability to answer-space drift and token-space drift remains in case of single-reward cluster-voting and self-certainty.

The \emph{multi-reward} setting without KL-Cov is more stable than either single reward, in line with the complementary structure of the two signals. Empirically, this complementarity broadens the stable training window: GSM8K remains in the $65$--$73\%$ range through step~200, and MATH500 stays above $46\%$ through step~180. 
Nevertheless, the run eventually collapses, with GSM8K falling from $56.6\%$ to $1.1\%$ between steps~240 and~260. Thus, multi-reward training delays degeneration but does not completely prevent it, motivating the need for KL-Cov regularization.

\begin{figure}[t]
\centering
\includegraphics[width=\textwidth]{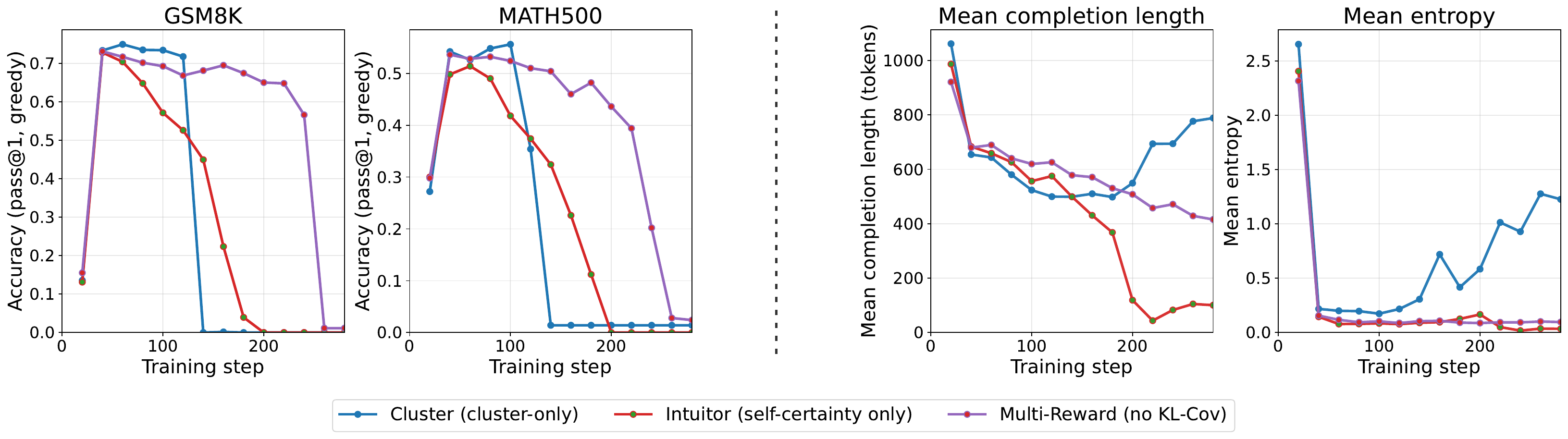}
\caption{Single- \& multi-reward analysis (no KL-Cov) on Qwen2.5-1.5B. Left: GSM8K and MATH500 pass@1. Right: training-time mean completion length and mean entropy. Cluster-only (blue) shows answer-mode collapse (rising entropy, lengthening generations); INTUITOR (red) shows token-mode collapse (entropy in $[0.03, 0.17]$, length crashing to ${\sim}100$ tokens); multi-reward without KL-Cov (purple) blocks both exploits but eventually collapses by step~260. {The answer-mode exploit is format-dependent: it requires an unbounded answer space, and the same multi-reward run retains $64\%$ of peak on ten-option MMLU-Pro (Table~\ref{tab:stability}).}}
\label{fig:single_multi_reward}
\end{figure}

\paragraph{KL-Cov coefficient sweep.}
We sweep $\beta_{\text{cov}} \in \{0.0005, 0.05, 0.1\}$ for the full multi-reward objective. Figure~\ref{fig:klcov_ablation} shows that all three settings produce tightly clustered trajectories, keeping completion length within $1000$--$1300$ tokens and maintaining stable math accuracy through step~280. By contrast, removing KL-Cov causes collapse around step~240, with short template-like completions and math accuracy dropping to ${\sim}1$--$2\%$. Additional results are provided in Table~\ref{tab:klcov_sweep} in Appendix~\ref{app:klcov_sweep}.
Overall, KL-Cov behaves as a robust stabilizer rather than a highly sensitive regularizer, and our default value $\beta_{\text{cov}}=0.05$ does not require per-model retuning.

\begin{figure}[t]
\centering
\includegraphics[width=.92\textwidth]{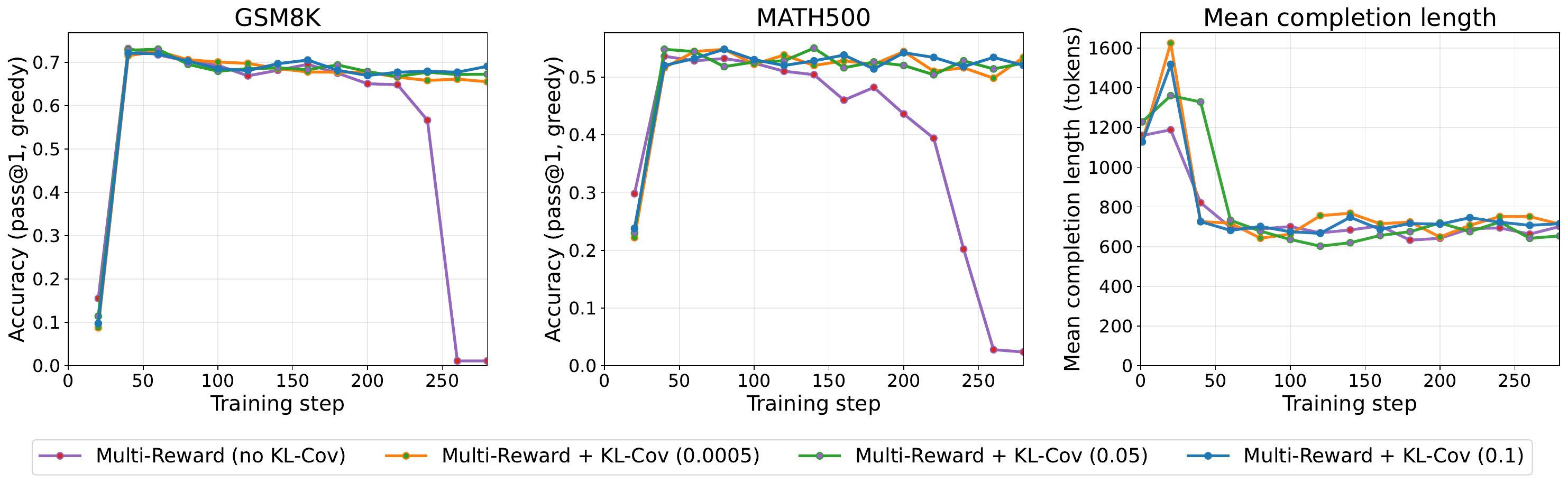}
\caption{KL-Cov $\beta_{\text{cov}}$ sweep on the full multi-reward objective (Qwen2.5-1.5B). All three sweeps of ($\beta_{\text{cov}} \in \{0.0005, 0.05, 0.1\}$) maintain stable accuracy and completion length through step~280 whereas no-KL-Cov baseline (purple) collapses early.}
\label{fig:klcov_ablation}
\end{figure}

%% ============================================================
%% 6. DISCUSSION
%% ============================================================
\section{Discussion}
\label{sec:discussion}
Learning without external verifiable rewards through RL post-training methods is crucial for domains with scarce, hard-to-verify, or unavailable gold-standard labels~\citep{ttrl, he2026unsupervised, intuitor, empo}. Our experiments show that single-reward RLIF approaches (cluster-voting, self-certainty) fail in complementary directions (Figure~\ref{fig:single_multi_reward}): cluster voting ignores the reasoning path, causing reasoning space drift  while concentrating answers into one cluster; self-certainty ignores answer correctness, so it can simply become deterministic. Combining them mitigates both, since each constrains what the other leaves underspecified. {The answer-space exploit is also \emph{format-dependent}: multi-reward without KL-Cov crashes on free-form GSM8K/MATH500 yet keeps $64\%$ of peak on ten-option MMLU-Pro (constrained answer space) (Table~\ref{tab:stability}; Appendix~\ref{app:qwen3}).} Multi-objective reward design assumes signals capture distinct but incomplete aspects of desired behavior and can be additively combined~\citep{intuitor}. Our work provides the first empirical demonstration of this in long horizon reasoning training. However, combining complementary signals alone is insufficient, and KL-Cov remains necessary to stabilize the residual dynamics. {It can also be augmented with supervised RLVR methods; our experiment GRPO-GT+KL-Cov shows a modest gain (Appendix~\ref{app:grpo_klcov}), but is decisive only in the unsupervised regime, where nothing external bounds the damage once token-space entropy collapses.}

\paragraph{Limitations.}
Our method introduces hyperparameters beyond standard GRPO{:} the top-$k$ covariance fraction, $\beta_{\text{cov}}$, and the channel weights $\alpha,\beta$. Figure~\ref{fig:klcov_ablation} shows that performance is largely insensitive to $\beta_{\text{cov}}$ across a $200{\times}$ range, but we do not study $\alpha/\beta$ and use the symmetric default $0.5$. Also, the cluster reward channel needs an answer-equivalence relation: exact for math and code; semantic clustering for open-ended reasoning and factual QA (NLI-based~\citep{semantic_uncertainty, empo}; not evaluated here). For multi-valued tasks, rubric-based equivalence is a natural direction. In addition, the method needs a base model already producing some correct rollouts, and we separate targeted KL-Cov from a global entropy bonus structurally, not empirically. Finally, our experimental evidence supports collapse \emph{resistance} over the long horizons, scales, and priors, not universal immunity, and each trajectory is a single run without seed repeats.
% ============================================================
%% 7. CONCLUSION
%% ============================================================

\section{Conclusion}
\label{sec:conclusion}

This paper introduces a multi-reward framework with KL-Cov regularization for scaling reinforcement learning training in LLMs without any external verifiers (RLIF). Our empirical results show that combining two complementary intrinsic signals mitigates the distinct degeneracies that these signals induce alone. KL-Cov further stabilizes optimization and delays late-stage collapse by targeting the small subset of tokens responsible for disproportionate entropy reduction. Across mathematical reasoning and coding benchmarks, our method performs close to supervised RLVR, such as GRPO, without gold-standard solutions, positioning multi-reward RLIF as a scalable alternative to externally supervised reinforcement learning. {For future work, two directions follow naturally: a genuinely online evaluation with streaming queries, which is the setting pure RLIF is ultimately for, and an extension to open-ended reasoning, where an advanced LLM-as-a-judge is now a practical, if compute-intensive, source of answer equivalence.}

\begin{ack}
\textbf{Funding in direct support of this work.} 
This work is partially supported by the Defense Advanced Research Projects Agency (DARPA) under Grant AI-CRAFT AWD16069.
Compute for all training and evaluation runs was provided by the Swiss National Supercomputing Centre (CSCS) under project ID a168 on Alps, as part of the ``Swiss AI Initiative''. No other third-party funding or support was received for this work.

\textbf{Competing interests.} The authors declare no financial relationships with entities that could be perceived to influence the submitted work, beyond the primary affiliations listed with the author names.
\end{ack}

\bibliographystyle{plainnat}
\bibliography{references}

@misc{gao2022scalinglawsrewardmodel,
      title={Scaling Laws for Reward Model Overoptimization}, 
      author={Leo Gao and John Schulman and Jacob Hilton},
      year={2022},
      eprint={2210.10760},
      archivePrefix={arXiv},
      primaryClass={cs.LG},
      url={https://arxiv.org/abs/2210.10760}, 
}

@article{deepseek_r1,
  title={{DeepSeek-R1}: Incentivizing Reasoning Capability in {LLMs} via Reinforcement Learning},
  author={{DeepSeek-AI}},
  journal={arXiv preprint arXiv:2501.12948},
  year={2025}
}

@article{mimo,
  title={{MiMo}: Unlocking the Reasoning Potential of Language Model -- From Pretraining to Posttraining},
  author={{Xiaomi LLM-Core Team}},
  journal={arXiv preprint arXiv:2505.07608},
  year={2025}
}

@article{intuitor,
  title={Learning to Reason without External Rewards},
  author={Zhao, Xuandong and Kang, Zhewei and Feng, Aosong and Levine, Sergey and Song, Dawn},
  journal={arXiv preprint arXiv:2505.19590},
  year={2025}
}

@article{empo,
  title={Right Question is Already Half the Answer: Fully Unsupervised {LLM} Reasoning Incentivization},
  author={Zhang, Qiang and Wu, Hao and Zhang, Changsheng and Zhao, Peilin and Bian, Yatao},
  journal={arXiv preprint arXiv:2504.05812},
  year={2025}
}

@article{entropy_effectiveness,
  title={The Unreasonable Effectiveness of Entropy Minimization in {LLM} Reasoning},
  author={Agarwal, Shivam and Zhang, Zimin and Yuan, Lifan and Han, Jiawei and Peng, Hao},
  journal={arXiv preprint arXiv:2505.15134},
  year={2025}
}

@article{srt,
  title={Can Large Reasoning Models Self-Train?},
  author={Shafayat, Sheikh and Tajwar, Fahim and Salakhutdinov, Ruslan and Schneider, Jeff and Zanette, Andrea},
  journal={arXiv preprint arXiv:2505.21444},
  year={2025}
}

@article{cui2025entropy,
  title={The Entropy Mechanism of Reinforcement Learning for Reasoning Language Models},
  author={Cui, Ganqu and Zhang, Yuchen and Chen, Jiacheng and Yuan, Lifan and Wang, Zhi and Zuo, Yuxin and Li, Haozhan and Fan, Yuchen and Chen, Huayu and Chen, Weize and Liu, Zhiyuan and Peng, Hao and Bai, Lei and Ouyang, Wanli and Cheng, Yu and Zhou, Bowen and Ding, Ning},
  journal={arXiv preprint arXiv:2505.22617},
  year={2025}
}

@article{nfl,
  title={No Free Lunch: Rethinking Internal Feedback for {LLM} Reasoning},
  author={Zhang, Yanzhi and Zhang, Zhaoxi and Guan, Haoxiang and Cheng, Yilin and Duan, Yitong and Wang, Chen and Wang, Yue and Zheng, Shuxin and He, Jiyan},
  journal={arXiv preprint arXiv:2506.17219},
  year={2025}
}

@article{gdpo,
  title={{GDPO}: Group Reward-Decoupled Normalization Policy Optimization for Multi-Reward {RL} Optimization},
  author={Liu, Shih-Yang and Dong, Xin and Lu, Ximing and Diao, Shizhe and Belcak, Peter and Liu, Mingjie and Chen, Min-Hung and Yin, Hongxu and Wang, Yu-Chiang Frank and Cheng, Kwang-Ting and Choi, Yejin and Kautz, Jan and Molchanov, Pavlo},
  journal={arXiv preprint arXiv:2601.05242},
  year={2026}
}

@article{grpo,
  title={{DeepSeekMath}: Pushing the Limits of Mathematical Reasoning in Open Language Models},
  author={Shao, Zhihong and Wang, Peiyi and Zhu, Qihao and Xu, Runxin and Song, Junxiao and Zhang, Mingchuan and Li, Y. K. and Wu, Y. and Guo, Daya},
  journal={arXiv preprint arXiv:2402.03300},
  year={2024}
}

@inproceedings{absolute_zero,
  title={Absolute Zero: Reinforced Self-play Reasoning with Zero Data},
  author={Zhao, Andrew and Wu, Yiran and Yue, Yang and Wu, Tong and Xu, Quentin and Lin, Matthieu and Wang, Shenzhi and Wu, Qingyun and Zheng, Zilong and Huang, Gao},
  booktitle={Advances in Neural Information Processing Systems (NeurIPS)},
  year={2025}
}

@article{ttrl,
  title={{TTRL}: Test-Time Reinforcement Learning},
  author={Zuo, Yuxin and Zhang, Kaiyan and Sheng, Li and Qu, Shang and Cui, Ganqu and Zhu, Xuekai and Li, Haozhan and Zhang, Yuchen and Long, Xinwei and Hua, Ermo and Qi, Biqing and Sun, Youbang and Ma, Zhiyuan and Yuan, Lifan and Ding, Ning and Zhou, Bowen},
  journal={arXiv preprint arXiv:2504.16084},
  year={2025}
}

@inproceedings{exploration_exploitation,
  title={Exploration vs Exploitation: Rethinking {RLVR} through Clipping, Entropy, and Spurious Reward},
  author={Chen, Peter and Li, Xiaopeng and Li, Ziniu and Yin, Wotao and Chen, Xi and Lin, Tianyi},
  booktitle={International Conference on Learning Representations (ICLR)},
  year={2026}
}

@article{corewarding,
  title={Co-rewarding: Stable Self-supervised {RL} for Eliciting Reasoning in Large Language Models},
  author={Zhang, Zizhuo and Zhu, Jianing and Ge, Xinmu and Zhao, Zihua and Zhou, Zhanke and Li, Xuan and Feng, Xiao and Yao, Jiangchao and Han, Bo},
  journal={arXiv preprint arXiv:2508.00410},
  year={2025}
}

@article{maximizing_confidence,
  title={Maximizing Confidence Alone Improves Reasoning},
  author={Prabhudesai, Mihir and Chen, Lili and Ippoliti, Alex and Fragkiadaki, Katerina and Liu, Hao and Pathak, Deepak},
  journal={arXiv preprint arXiv:2505.22660},
  year={2025}
}

@article{verifree,
  title={Reinforcing General Reasoning without Verifiers},
  author={Zhou, Xiangxin and Liu, Zichen and Sims, Anya and Wang, Haonan and Pang, Tianyu and Li, Chongxuan and Wang, Liang and Lin, Min and Du, Chao},
  journal={arXiv preprint arXiv:2505.21493},
  year={2025}
}

@article{rzero,
  title={{R-Zero}: Self-Evolving Reasoning {LLM} from Zero Data},
  author={Huang, Chengsong and Yu, Wenhao and Wang, Xiaoyang and Zhang, Hongming and Li, Zongxia and Li, Ruosen and Huang, Jiaxin and Mi, Haitao and Yu, Dong},
  journal={arXiv preprint arXiv:2508.05004},
  year={2025}
}

@inproceedings{he2026unsupervised,
  title={How Far Can Unsupervised {RLVR} Scale {LLM} Training?},
  author={He, Bingxiang and Zuo, Yuxin and Liu, Zeyuan and Zhao, Shangziqi and Fu, Zixuan and Yang, Junlin and Qian, Cheng and Zhang, Kaiyan and Fan, Yuchen and Cui, Ganqu and Chen, Xiusi and Sun, Youbang and Lv, Xingtai and Zhu, Xuekai and Sheng, Li and Li, Ran and Gao, Huan-ang and Zhang, Yuchen and Zhou, Bowen and Liu, Zhiyuan and Ding, Ning},
  booktitle={International Conference on Learning Representations (ICLR)},
  year={2026}
}

@article{ssr_zero,
  title={{SSR-Zero}: Simple Self-Rewarding Reinforcement Learning for Machine Translation},
  author={Yang, Wenjie and Zheng, Mao and Song, Mingyang and Li, Zheng and Wang, Sitong},
  journal={arXiv preprint arXiv:2505.16637},
  year={2025}
}

@inproceedings{semantic_uncertainty,
  title={Semantic Uncertainty: Linguistic Invariances for Uncertainty Estimation in Natural Language Generation},
  author={Kuhn, Lorenz and Gal, Yarin and Farquhar, Sebastian},
  booktitle={International Conference on Learning Representations (ICLR)},
  year={2023}
}

@inproceedings{livecodebench,
  title={{LiveCodeBench}: Holistic and Contamination Free Evaluation of Large Language Models for Code},
  author={Jain, Naman and Han, King and Gu, Alex and Li, Wen-Ding and Yan, Fanjia and Zhang, Tianjun and Wang, Sida and Solar-Lezama, Armando and Sen, Koushik and Stoica, Ion},
  booktitle={International Conference on Learning Representations (ICLR)},
  year={2025},
  url={https://arxiv.org/abs/2403.07974}
}

@inproceedings{cruxeval,
  title={{CRUXEval}: A Benchmark for Code Reasoning, Understanding and Execution},
  author={Gu, Alex and Rozi{\`e}re, Baptiste and Leather, Hugh and Solar-Lezama, Armando and Synnaeve, Gabriel and Wang, Sida},
  booktitle={International Conference on Machine Learning (ICML)},
  year={2024},
  url={https://arxiv.org/abs/2401.03065}
}

@misc{lighteval,
  title={Lighteval: A lightweight framework for {LLM} evaluation},
  author={Habib, Nathan and Fourrier, Cl{\'e}mentine and Kydl{\'\i}{\v{c}}ek, Hynek and Wolf, Thomas and Tunstall, Lewis},
  year={2023},
  howpublished={\url{https://github.com/huggingface/lighteval}}
}

@misc{zeroeval,
  title={{ZeroEval}: A Unified Framework for Evaluating Language Models},
  author={Lin, Yuchen},
  year={2024},
  howpublished={\url{https://github.com/yuchenlin/ZeroEval}}
}

@article{hendrycksmath2021,
    title={Measuring Mathematical Problem Solving With the MATH Dataset},
    author={Dan Hendrycks
    and Collin Burns
    and Saurav Kadavath
    and Akul Arora
    and Steven Basart
    and Eric Tang
    and Dawn Song
    and Jacob Steinhardt},
    journal={arXiv preprint arXiv:2103.03874},
    year={2021}
}

@article{qwen2.5,
  title={{Qwen2.5} Technical Report},
  author={Yang, An and Yang, Baosong and Zhang, Beichen and Hui, Binyuan and Zheng, Bo and Yu, Bowen and Li, Chengyuan and Liu, Dayiheng and Huang, Fei and Wei, Haoran and others},
  journal={arXiv preprint arXiv:2412.15115},
  year={2024}
}

@article{cobbe2021gsm8k,
  title={Training Verifiers to Solve Math Word Problems},
  author={Cobbe, Karl and Kosaraju, Vineet and Bavarian, Mohammad and Chen, Mark and Jun, Heewoo and Kaiser, Lukasz and Plappert, Matthias and Tworek, Jerry and Hilton, Jacob and Nakano, Reiichiro and Hesse, Christopher and Schulman, John},
  journal={arXiv preprint arXiv:2110.14168},
  year={2021}
}

@inproceedings{wang2024mmlupro,
  title={{MMLU-Pro}: A More Robust and Challenging Multi-Task Language Understanding Benchmark},
  author={Wang, Yubo and Ma, Xueguang and Zhang, Ge and Ni, Yuansheng and Chandra, Abhranil and Guo, Shiguang and Ren, Weiming and Arulraj, Aaran and He, Xuan and Jiang, Ziyan and Li, Tianle and Ku, Max and Wang, Kai and Zhuang, Alex and Fan, Rongqi and Yue, Xiang and Chen, Wenhu},
  booktitle={Advances in Neural Information Processing Systems (NeurIPS)},
  year={2024}
}

@misc{vonwerra2020trl,
  title   = {{TRL: Transformers Reinforcement Learning}},
  author  = {von Werra, Leandro and Belkada, Younes and Tunstall, Lewis and Beeching, Edward and Thrush, Tristan and Lambert, Nathan and Huang, Shengyi and Rasul, Kashif and Gallouédec, Quentin},
  license = {Apache-2.0},
  url     = {https://github.com/huggingface/trl},
  year    = {2020}
}

@article{chen2021evaluating,
  title={Evaluating Large Language Models Trained on Code},
  author={Chen, Mark and Tworek, Jerry and Jun, Heewoo and Yuan, Qiming and de Oliveira Pinto, Henrique Ponde and Kaplan, Jared and Edwards, Harri and Burda, Yuri and Joseph, Nicholas and Brockman, Greg and others},
  journal={arXiv preprint arXiv:2107.03374},
  year={2021}
}

%%%%%%%%%%%%%%%%%%%%%%%%%%%%%%%%%%%%%%%%%%%%%%%%%%%%%%%%%%%%

\appendix

\section{Performance on AIME}
\label{app:aime}
AIME~2024/2025 are at the 30-problem noise floor at the 1.5B scale (one problem ${\approx}3.3$ pp), so 1.5B numbers should be read as near-zero indicators only. On 3B, multi-reward (without KL-Cov) ties supervised GRPO-GT and INTUITOR ($6.7\%$ on AIME 2024, $3.3\%$ on AIME 2025), while multi-reward~+~KL-Cov drops to $3.3\%$ / $0.0\%$.

\begin{table}[ht]
\caption{Pass@1 (\%, greedy) on AIME 2024 and AIME 2025 at the step-160 checkpoint. Best \emph{unsupervised} value per column in \textbf{bold} (ties bolded together).}
\label{tab:aime_results}
\centering
\small
\begin{tabular}{llcc}
\toprule
\textbf{Model} & \textbf{Method} & \textbf{AIME 2024} & \textbf{AIME 2025} \\
\midrule
\multirow{4}{*}{Qwen2.5-1.5B}
 & GRPO with ground-truth reward                & 0.0 & 3.3 \\
 & INTUITOR                                     & 0.0 & 0.0 \\
 & Multi-reward (ours)                          & 0.0 & 0.0 \\
 & Multi-reward + KL-Cov (ours) $\beta_{cov}{=}0.05$ & 0.0 & 0.0 \\
\midrule
\multirow{4}{*}{Qwen2.5-3B}
 & GRPO with ground-truth reward                & 6.7 & 3.3 \\
 & INTUITOR                                     & \textbf{6.7} & \textbf{3.3} \\
 & Multi-reward (ours)                          & \textbf{6.7} & \textbf{3.3} \\
 & Multi-reward + KL-Cov (ours) $\beta_{cov}{=}0.05$ & 3.3 & 0.0 \\
\bottomrule
\end{tabular}
\end{table}

\section{Scaling to Larger Models}
\label{app:7b}

We repeat the trajectory study on Qwen2.5-7B for supervised \emph{GRPO-GT}, single-signal \emph{INTUITOR}, and \emph{Multi-reward + KL-Cov} ($\beta_{\text{cov}}{=}0.05$). Hyperparameters match the 3B configuration (Appendix~\ref{app:hyperparams}); only the per-device batch and grad-accumulation layout differ to fit Qwen2.5-7B on $4{\times}$GH200. Each run is evaluated every $20$ steps through step~$240$ (the longest common horizon under our $12$\,h walltime).

\paragraph{Greedy ($T{=}0$) findings} (Figure~\ref{fig:trajectory_7b}, Tables~\ref{tab:trajectory_7b}--\ref{tab:peak_7b}). \textbf{(1) INTUITOR's collapse is sharper at 7B:} pass@1 holds near base-model accuracy through step~$80$, then drops to $0.0$\,\% on GSM8K and MATH500 from step~$100$ onward (AIME stays at noise floor $\le 3.3$\,\%). The single-signal reward thus reaches a global degenerate {fixed point} in well under one epoch. \textbf{(2) Multi-reward + KL-Cov tracks GRPO-GT:} per-step gap averages $1.8$\,pp on GSM8K and $2.1$\,pp on MATH500 (max $3.9$\,/\,$4.4$\,pp; final $0.2$\,/\,$0.8$\,pp at step~$240$). No collapse phase is observed. \textbf{(3) AIME peaks tie:} multi-reward + KL-Cov $13.3$\,/\,$16.7$\,\% on 2024/2025 vs.\ GRPO-GT $16.7$\,/\,$13.3$\,\% (one-problem difference in opposite directions; AIME has $30$ problems per year, so this is a sanity check rather than a ranking).

\begin{figure}[ht]
\centering
\includegraphics[width=\textwidth]{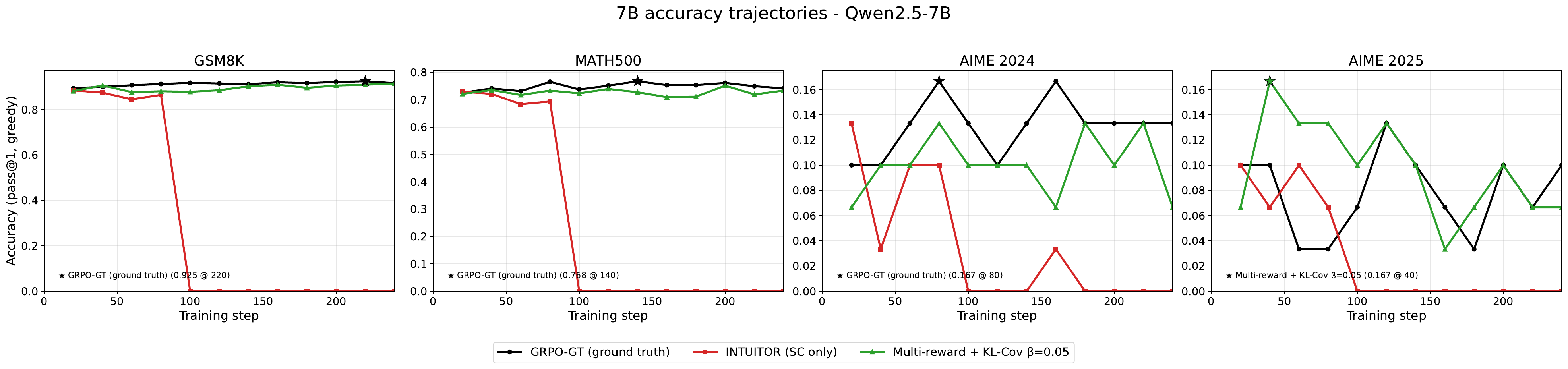}
\caption{Greedy pass@1 over training on Qwen2.5-7B. \emph{Multi-reward + KL-Cov} tracks supervised \emph{GRPO-GT} within a few pp; \emph{INTUITOR} collapses from step~$100$ (math $\to 0$\,\%, AIME at noise floor). Stars: per-panel global peaks.}
\label{fig:trajectory_7b}
\end{figure}

\begin{table}[ht]
\caption{Qwen2.5-7B step-$240$ snapshot (\%). \emph{Greedy}: $T{=}0$ pass@1. \emph{avg-P@1}, \emph{pass@$K$}: mean and unbiased coverage~\citep{chen2021evaluating} over $N$ samples at $T{=}1$ ($N{=}8$/$16$/$32$ for GSM8K/MATH500/AIME). Per-step trajectories: Figures~\ref{fig:trajectory_7b} and~\ref{fig:passk_combined_7b}. \textbf{Bold}: best unsupervised per column.}
\label{tab:trajectory_7b}
\centering
\small
\setlength{\tabcolsep}{3pt}
\resizebox{\textwidth}{!}{%
\begin{tabular}{lccccccccccccc}
\toprule
& \multicolumn{3}{c}{\textbf{GSM8K}} & \multicolumn{3}{c}{\textbf{MATH500}} & \multicolumn{3}{c}{\textbf{AIME 2024}} & \multicolumn{3}{c}{\textbf{AIME 2025}} \\
\cmidrule(lr){2-4}\cmidrule(lr){5-7}\cmidrule(lr){8-10}\cmidrule(lr){11-13}
\textbf{Method}
 & Greedy & avg-P@1 & pass@$K$
 & Greedy & avg-P@1 & pass@$K$
 & Greedy & avg-P@1 & pass@$K$
 & Greedy & avg-P@1 & pass@$K$ \\
\midrule
GRPO-GT (supervised)
 & 91.7 & 89.1 & 97.4
 & 74.2 & 72.0 & 90.6
 & 13.3 & 14.2 & 40.0
 & 10.0 &  4.6 & 30.0 \\
INTUITOR
 &  0.0 &  0.0 &  0.0
 &  0.0 &  0.0 &  0.0
 &  0.0 &  0.0 &  0.0
 &  0.0 &  0.0 &  0.0 \\
Multi-reward + KL-Cov (ours) $\beta_{cov}{=}0.05$ (ours)
 & \textbf{91.5} & \textbf{89.9} & \textbf{96.7}
 & \textbf{73.4} & \textbf{71.6} & \textbf{88.2}
 & \textbf{6.7}  & \textbf{9.1}  & \textbf{36.7}
 & \textbf{6.7}  & \textbf{6.3}  & \textbf{30.0} \\
\bottomrule
\end{tabular}%
}
\end{table}

\begin{table}[ht]
\caption{Peak values during training on GSM8K and MATH500 (\%), step in parentheses (earliest if tied). Greedy: every $20$ steps; $T{=}1$: steps $\{40,80,120,160,200,240\}$. AIME is omitted as $T{=}1$ was measured only at step~$240$ (see Table~\ref{tab:trajectory_7b}). INTUITOR's peak at step~$20$ reflects near-base-model accuracy before single-signal collapse. \textbf{Bold}: best unsupervised per column.}
\label{tab:peak_7b}
\centering
\small
\setlength{\tabcolsep}{4pt}
\resizebox{\textwidth}{!}{%
\begin{tabular}{lcccccc}
\toprule
& \multicolumn{3}{c}{\textbf{GSM8K}} & \multicolumn{3}{c}{\textbf{MATH500}} \\
\cmidrule(lr){2-4}\cmidrule(lr){5-7}
\textbf{Method}
 & Greedy & avg-P@1 & pass@$K$
 & Greedy & avg-P@1 & pass@$K$ \\
\midrule
GRPO-GT (sup.)
 & 92.5\,(220) & 89.2\,(200) & 97.6\,(80)
 & 76.8\,(140) & 72.7\,(200) & 91.6\,(40) \\
INTUITOR
 & 88.5\,(20)  & 85.9\,(80)  & 96.7\,(40)
 & 73.0\,(20)  & 68.3\,(40)  & 89.8\,(40) \\
Multi-reward + KL-Cov (ours) $\beta_{cov}{=}0.05$ (ours)
 & \textbf{91.5}\,(240) & \textbf{89.9}\,(240) & \textbf{97.1}\,(40)
 & \textbf{75.2}\,(200) & \textbf{71.6}\,(240) & \textbf{90.2}\,(40) \\
\bottomrule
\end{tabular}%
}
\end{table}

\subsection{Temperature-\texorpdfstring{$1$}{1} sampling}
\label{sec:7b_passk}

To rule out greedy-decoding artifacts, we re-evaluate the same checkpoints under the avg-Pass@1 convention~\citep{deepseek_r1, ttrl, chen2021evaluating}: $N$ samples per problem at $T{=}1$ ($N{=}8$ for GSM8K, $N{=}16$ for MATH500), averaged across samples, and report the unbiased pass@$K_{\max}$ coverage at $k{=}N$. Two observations (Figure~\ref{fig:passk_combined_7b}) strengthen the greedy story:
\textbf{(i) INTUITOR's collapse is total, not mode-selection.} From step~$120$ onward both avg-Pass@1 and pass@$K_{\max}$ are $0.0$\,\% on GSM8K and MATH500, no correct sample on any test problem in $K_{\max}$ tries; the policy has lost the math-reasoning manifold rather than concentrating on the wrong mode.
\textbf{(ii) Multi-reward + KL-Cov matches GRPO-GT under sampling.} GSM8K avg-Pass@1 at step~$240$: $89.9$\,\% (ours) vs.\ $89.1$\,\% (GRPO-GT, $+0.8$\,pp with no ground-truth reward); MATH500 within $1.4$\,pp at every step; pass@$16$ on MATH500 at step~$120$ is $90.2$\,\% (essentially full coverage), confirming the policy retains breadth rather than collapsing to a sharp mode.

\begin{figure}[ht]
\centering
\includegraphics[width=\textwidth]{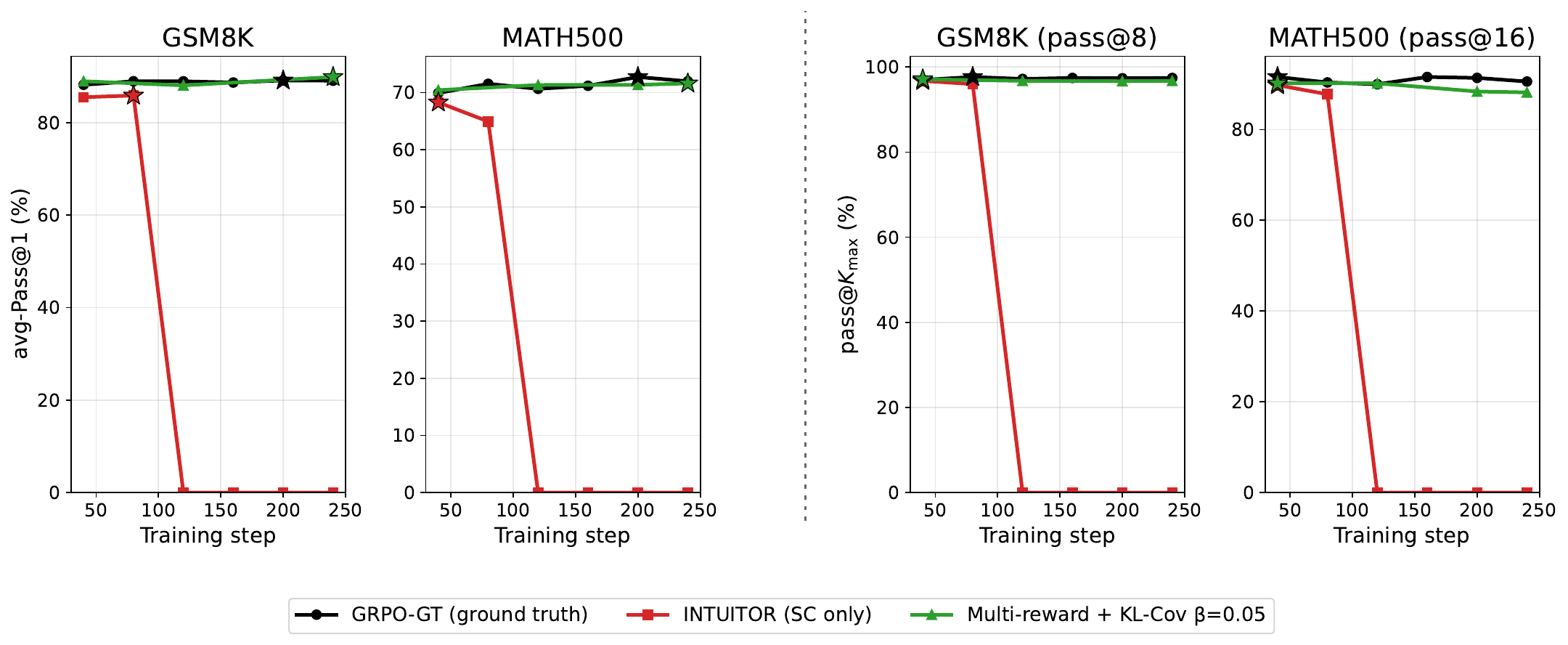}
\caption{Qwen2.5-7B $T{=}1.0$ sampling trajectories. \textbf{Left:} avg-Pass@1 on GSM8K ($N{=}8$) and MATH500 ($N{=}16$). \textbf{Right:} pass@$K_{\max}$ coverage ({\citet{chen2021evaluating}} unbiased estimator at $k{=}N$). Multi-reward + KL-Cov tracks GRPO-GT on both metrics ($\le 1.5$\,/\,$2.5$\,pp final-step gap); INTUITOR drops to $0$\,\% on both from step~$120$. Stars: per-method peaks. Multi-reward + KL-Cov steps $80$ and $160$ were not retained on disk.}
\label{fig:passk_combined_7b}
\end{figure}

\section{Newer-Generation Backbone: Qwen3-1.7B-Base}
\label{app:qwen3}

To test whether the collapse-resistance mechanism transfers across model priors rather than only across scales, we repeat the trajectory study on \textbf{Qwen3-1.7B-Base}, a newer-generation base model with a substantially different pretraining recipe than Qwen2.5, for supervised GRPO-GT, single-signal INTUITOR, and multi-reward + KL-Cov ($\beta_{\text{cov}}{=}0.02$). The parameter scale is deliberately matched to our 1.5B suite so that the prior change is isolated from the scale change (Appendix~\ref{app:7b} covers scale). The optimization recipe is unchanged (LR $3{\times}10^{-6}$, $\gamma{=}0.005$, $\epsilon{=}0.2$, $T{=}1.0$, $G{=}7$, max completion $3072$; effective batch $882$ completions via per-device batch $7$ and gradient accumulation $42$ on three 48\,GB training GPUs plus one for vLLM generation, with $\mu{=}2$ PPO epochs per rollout batch). Each run trains for $200$ steps and every checkpoint (every $20$ steps) is evaluated greedily ($T{=}0$, pass@1) on GSM8K and MATH500; MMLU-Pro is evaluated at step~$200$.

\paragraph{Findings} (Figure~\ref{fig:trajectory_a2}, Tables~\ref{tab:snapshot_a2}--\ref{tab:peak_a2}). \textbf{(1) Single-signal degradation transfers to the newer prior, as sustained decay rather than abrupt collapse:} INTUITOR peaks at step~$40$ ($76.0$\,\% GSM8K, $62.6$\,\% MATH500), then declines steadily, GSM8K falls $19.4$\,pp to a trough of $56.6$\,\% at step~$120$, and from step~$180$ MATH500 drops \emph{below the base model} ($49.2$\,\% vs.\ $52.8$\,\%), ending at $61.5$\,/\,$51.0$\,\%. The failure mode is thus present under the Qwen3 prior, though it manifests as monotone degradation rather than the global fixed-point collapse observed on Qwen2.5-7B (Appendix~\ref{app:7b}). \textbf{(2) Multi-reward + KL-Cov is stable across the horizon:} no decline phase is observed; the method retains $97.8$\,\% (GSM8K) and $95.8$\,\% (MATH500) of its peak at step~$200$. \textbf{(3) It tracks and early on exceeds supervised GRPO-GT:} the mean per-step gap is $2.3$\,pp on GSM8K and $0.9$\,pp on MATH500, and our label-free method leads GRPO-GT over steps $20$--$80$ on MATH500 (peak $66.0$ vs.\ $66.2$ overall). On MMLU-Pro at step~$200$ all three methods improve over the base model ($36.5$\,\%); consistent with the format-dependence analysis of Section~\ref{sec:entropy_analysis}, the constrained answer space also shields INTUITOR from its free-form degradation ($38.8$\,\% vs.\ ours $38.3$\,\%, GRPO-GT $39.8$\,\%).

\paragraph{Statistical significance}Wilson 95\% CIs at step~200 (GSM8K $n{=}1319$, MATH500 $n{=}500$, MMLU-Pro $n{=}2000$): ours $80.1\,[77.8,82.1]$ / $63.2\,[58.9,67.3]$ / $38.3\,[36.2,40.5]$; INTUITOR $61.5\,[58.8,64.1]$ / $51.0\,[46.6,55.4]$ / $38.8\,[36.6,40.9]$; GRPO-GT $83.4\,[81.3,85.3]$ / $64.8\,[60.5,68.9]$ / $39.8\,[37.6,41.9]$. Exact McNemar tests on paired per-item outcomes: ours vs.\ INTUITOR is decisive on the free-form benchmarks (GSM8K $p{=}3.0{\times}10^{-42}$, MATH500 $p{=}1.9{\times}10^{-8}$) but not significant on ten-option MMLU-Pro ($p{=}0.67$), statistically confirming the format-dependence claim of Section~\ref{sec:entropy_analysis}. Against supervised GRPO-GT, ours is statistically indistinguishable on MATH500 ($p{=}0.48$) and MMLU-Pro ($p{=}0.14$); the residual GSM8K gap ($3.3$\,pp, $p{=}0.0018$) is real but small.

\begin{figure}[ht]
\centering
\includegraphics[width=\textwidth]{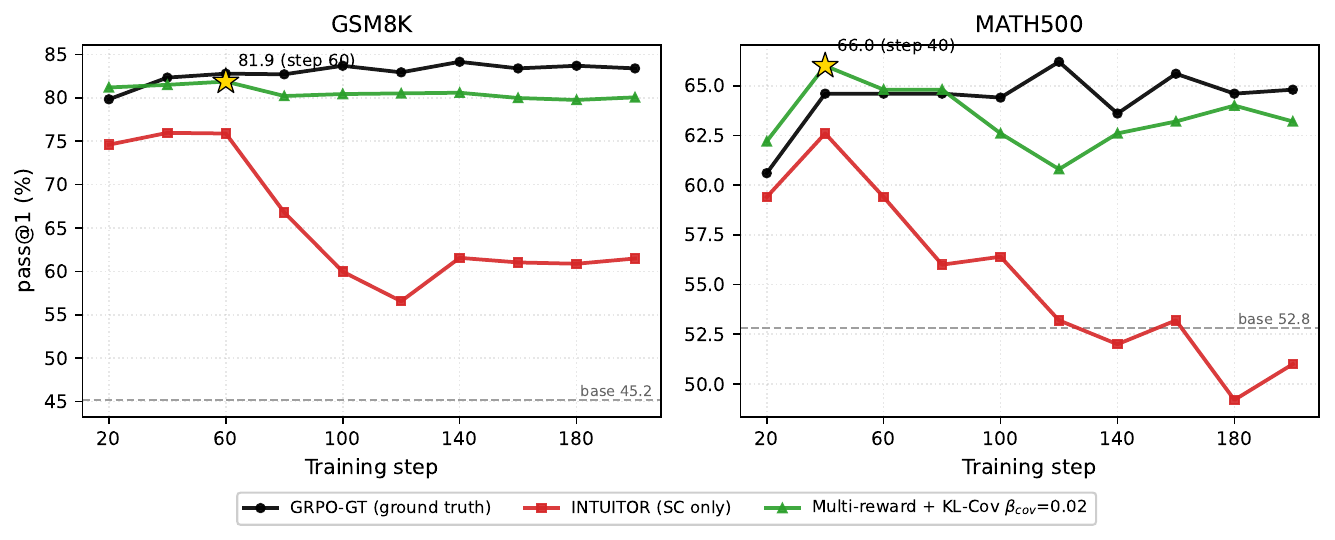}
\caption{Greedy pass@1 over training on Qwen3-1.7B-Base. \emph{Multi-reward + KL-Cov} is stable across the full 200-step horizon and tracks supervised \emph{GRPO-GT} within a few pp; \emph{INTUITOR} peaks at step~$40$ and degrades thereafter, falling below the base model (gray dashed line) on MATH500 from step~$180$. Stars: per-panel peaks among unsupervised methods.}
\label{fig:trajectory_a2}
\end{figure}

\begin{table}[ht]
\caption{Qwen3-1.7B-Base step-$200$ snapshot (greedy pass@1, \%). \textbf{Bold}: best unsupervised per column. Per-step trajectories: Figure~\ref{fig:trajectory_a2}.}
\label{tab:snapshot_a2}
\centering
\small
\begin{tabular}{lccc}
\toprule
\textbf{Method} & \textbf{GSM8K} & \textbf{MATH500} & \textbf{MMLU-Pro} \\
\midrule
Qwen3-1.7B-Base (no RL) & 45.2 & 52.8 & 36.5 \\
GRPO-GT (supervised) & 83.4 & 64.8 & 39.8 \\
INTUITOR & 61.5 & 51.0 & \textbf{38.8} \\
Multi-reward + KL-Cov (ours) & \textbf{80.1} & \textbf{63.2} & 38.3 \\
\bottomrule
\end{tabular}
\end{table}

\begin{table}[ht]
\caption{Peak, end-of-horizon, and retention on Qwen3-1.7B-Base (greedy pass@1, \%), peak step in parentheses. Retention $=$ step-$200$ accuracy $/$ peak accuracy. \textbf{Bold}: best unsupervised per column.}
\label{tab:peak_a2}
\centering
\small
\setlength{\tabcolsep}{4pt}
\begin{tabular}{lcccccc}
\toprule
& \multicolumn{3}{c}{\textbf{GSM8K}} & \multicolumn{3}{c}{\textbf{MATH500}} \\
\cmidrule(lr){2-4}\cmidrule(lr){5-7}
\textbf{Method} & Peak (step) & Step 200 & Retention & Peak (step) & Step 200 & Retention \\
\midrule
GRPO-GT (supervised) & 84.2\,(140) & 83.4 & 99.0\,\% & 66.2\,(120) & 64.8 & 97.9\,\% \\
INTUITOR & 76.0\,(40) & 61.5 & 80.9\,\% & 62.6\,(40) & 51.0 & 81.5\,\% \\
Multi-reward + KL-Cov (ours) & \textbf{81.9}\,(60) & \textbf{80.1} & \textbf{97.8\,\%} & \textbf{66.0}\,(40) & \textbf{63.2} & \textbf{95.8\,\%} \\
\bottomrule
\end{tabular}
\end{table}

\section{Augmenting Supervised RLVR: GRPO-GT with and without KL-Cov}
\label{app:grpo_klcov}

The main results establish KL-Cov as a component of the multi-reward + KL-Cov objective. A natural question is whether it also transfers to the supervised setting, i.e., whether GRPO trained on ground-truth rewards benefits from the same targeted covariance regularizer. We therefore train two arms from Qwen2.5-1.5B under identical budgets: \textbf{GRPO-GT} (accuracy reward only) and \textbf{GRPO-GT + KL-Cov} ($\beta_{\text{cov}}{=}0.02$, top-$2\%$ covariance tokens, as in the INTUITOR$+$KL-Cov ablation of Appendix~\ref{app:hyperparams}). Both arms share the optimization recipe of our trajectory studies (LR $3{\times}10^{-6}$, $\gamma{=}0.005$, $\epsilon{=}0.2$, $T{=}1.0$, $G{=}7$, max completion $3072$), train for $200$ steps on MATH, and are evaluated greedily ($T{=}0$, pass@1) at every checkpoint (every $20$ steps) on GSM8K and MATH500. Under this zero-shot boxed protocol the base model scores $1.7$\,\% / $9.2$\,\% (GSM8K / MATH500), far below the plotted range, so no base reference line is drawn in Figure~\ref{fig:trajectory_a4}.

\paragraph{Findings} (Figure~\ref{fig:trajectory_a4}, Tables~\ref{tab:traj_a4}--\ref{tab:peak_a4}). \textbf{(1) KL-Cov transfers to the supervised objective as a drop-in regularizer:} it never destabilizes training and ends at or above the plain baseline on both benchmarks. \textbf{(2) On GSM8K it converts a plateau into continued improvement:} plain GRPO-GT reaches its peak ($74.3$\,\%) already at step~$40$ and oscillates below it thereafter, whereas the KL-Cov arm keeps improving through step~$140$ ($76.1$\,\%) and holds a consistent advantage over the entire final stretch: it is higher at all five checkpoints from step~$120$ onward (mean $+2.0$\,pp), ending at $75.7$ vs.\ $73.3$\,\%. \textbf{(3) On MATH500 the two arms are statistically comparable}, with a pattern consistent with the GSM8K one at smaller magnitude: the plain baseline peaks higher mid-training ($58.4$ at step~$80$) but decays to $54.8$\,\% ($93.8$\,\% retention), while the KL-Cov arm stays flat around $55$\,\% and ends slightly higher ($55.6$\,\%, $96.9$\,\% retention). \textbf{(4) The gains are deliberately modest and that is the expected outcome:} the gold-label reward already anchors training, so GRPO-GT does not collapse within this horizon and there is little instability for KL-Cov to remove. Its contribution in the supervised regime is late-stage entropy preservation, the same mechanism that, in the unsupervised regime where no gold signal anchors the policy, is the difference between stability and collapse (Section~\ref{sec:entropy_analysis}, Appendix~\ref{app:klcov_sweep}). The one visible cost is a slower first checkpoint on GSM8K ($57.8$ vs.\ $66.5$\,\% at step~$20$), as the regularizer moderates the earliest, largest updates; the gap closes by step~$40$.

\paragraph{Statistical significance}At step~200 the GSM8K advantage of the KL-Cov arm is significant at $\alpha{=}0.05$ under an exact McNemar test on paired per-item outcomes ($75.7$ vs.\ $73.3$\,\%, $p{=}0.030$; Wilson 95\% CIs $[73.3,77.9]$ vs.\ $[70.9,75.6]$, $n{=}1319$), while MATH500 is statistically comparable ($55.6$ vs.\ $54.8$\,\%, $p{=}0.74$, $n{=}500$), quantifying findings (2) and (3).

\begin{figure}[ht]
\centering
\includegraphics[width=\textwidth]{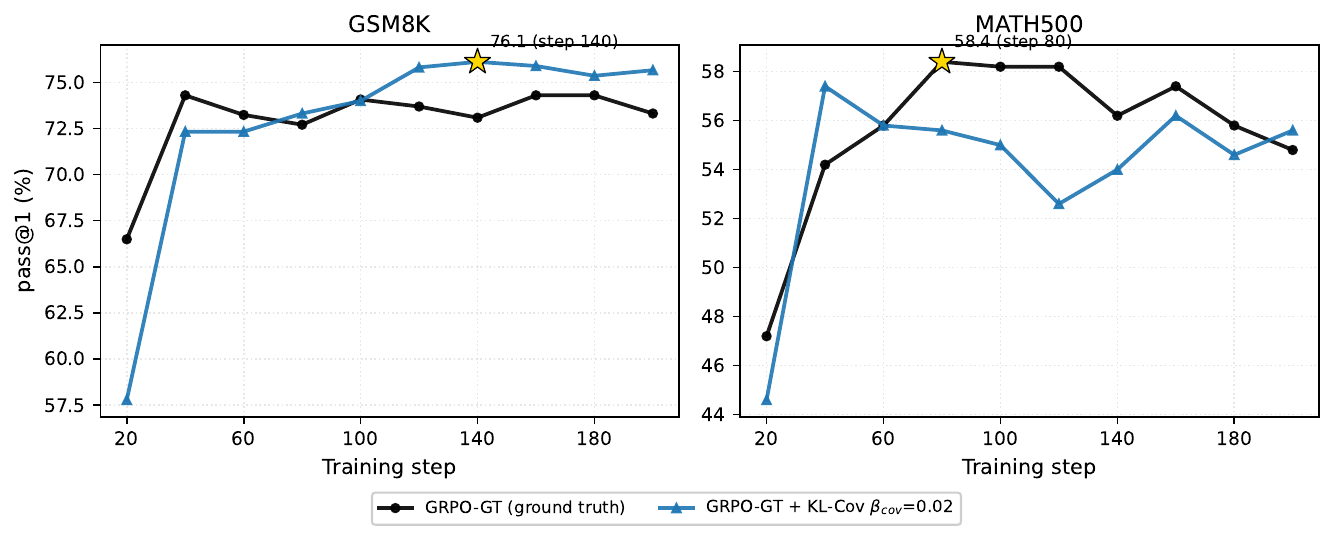}
\caption{Greedy pass@1 over training on Qwen2.5-1.5B for supervised \emph{GRPO-GT} and \emph{GRPO-GT + KL-Cov}. On GSM8K the plain baseline plateaus at step~$40$ while the KL-Cov arm keeps improving through step~$140$ and stays above the baseline over the final five checkpoints; on MATH500 the arms are comparable, with the KL-Cov arm retaining more of its peak. Stars: per-panel peak across both arms.}
\label{fig:trajectory_a4}
\end{figure}

\begin{table}[ht]
\caption{Per-checkpoint greedy pass@1 (\%) on Qwen2.5-1.5B for GRPO-GT with and without KL-Cov. \textbf{Bold}: better arm per column.}
\label{tab:traj_a4}
\centering
\small
\setlength{\tabcolsep}{4pt}
\begin{tabular}{llcccccccccc}
\toprule
& \textbf{Method} & \textbf{20} & \textbf{40} & \textbf{60} & \textbf{80} & \textbf{100} & \textbf{120} & \textbf{140} & \textbf{160} & \textbf{180} & \textbf{200} \\
\midrule
\multirow{2}{*}{GSM8K}
& GRPO-GT           & \textbf{66.5} & \textbf{74.3} & \textbf{73.2} & 72.7 & \textbf{74.1} & 73.7 & 73.1 & 74.3 & 74.3 & 73.3 \\
& \; + KL-Cov       & 57.8 & 72.3 & 72.3 & \textbf{73.3} & 74.0 & \textbf{75.8} & \textbf{76.1} & \textbf{75.9} & \textbf{75.4} & \textbf{75.7} \\
\midrule
\multirow{2}{*}{MATH500}
& GRPO-GT           & \textbf{47.2} & 54.2 & 55.8 & \textbf{58.4} & \textbf{58.2} & \textbf{58.2} & \textbf{56.2} & \textbf{57.4} & \textbf{55.8} & 54.8 \\
& \; + KL-Cov       & 44.6 & \textbf{57.4} & 55.8 & 55.6 & 55.0 & 52.6 & 54.0 & 56.2 & 54.6 & \textbf{55.6} \\
\bottomrule
\end{tabular}
\end{table}

\begin{table}[ht]
\caption{Peak, end-of-horizon, and retention on Qwen2.5-1.5B for GRPO-GT with and without KL-Cov (greedy pass@1, \%), peak step in parentheses. Retention $=$ step-$200$ accuracy $/$ peak accuracy. \textbf{Bold}: better arm per column.}
\label{tab:peak_a4}
\centering
\small
\setlength{\tabcolsep}{4pt}
\begin{tabular}{lcccccc}
\toprule
& \multicolumn{3}{c}{\textbf{GSM8K}} & \multicolumn{3}{c}{\textbf{MATH500}} \\
\cmidrule(lr){2-4}\cmidrule(lr){5-7}
\textbf{Method} & Peak (step) & Step 200 & Retention & Peak (step) & Step 200 & Retention \\
\midrule
GRPO-GT      & 74.3\,(40) & 73.3 & 98.7\,\% & \textbf{58.4}\,(80) & 54.8 & 93.8\,\% \\
\; + KL-Cov  & \textbf{76.1}\,(140) & \textbf{75.7} & \textbf{99.5\,\%} & 57.4\,(40) & \textbf{55.6} & \textbf{96.9\,\%} \\
\bottomrule
\end{tabular}
\end{table}

\section{Detailed stability analysis}
\label{app:stability}

This appendix accompanies Figure~\ref{fig:trajectory} (Section~\ref{sec:entropy_analysis}) with the per-method numerical breakdown referenced in the main text. Table~\ref{tab:stability} reports peak accuracy, peak step, collapse step, end-of-horizon accuracy, and retention for both Qwen2.5-1.5B and Qwen2.5-3B. Three observations follow from the figure together with this table; the AIME~2024/2025 far-OOD probe is reported separately in Appendix~\ref{app:aime}.

\begin{table}[ht]
\caption{\textbf{Stability under continued training} (greedy pass@1) for Qwen2.5-1.5B and Qwen2.5-3B. Each cell reports the metric on three benchmarks as \texttt{GSM8K\,/\,MATH500\,/\,MMLU-Pro}. \emph{Peak (\%)} is the best pass@1 reached during the trajectory, and \emph{Peak step} is the first checkpoint at which that peak is reached. \emph{Collapse step} is the first checkpoint \emph{after peak} at which accuracy drops below $50\%$ of peak (``---'' if never crossed by step 340). \emph{Step-340} is end-of-horizon accuracy and \emph{Retention} is step-340\,/\,peak.}
\label{tab:stability}
\centering
\small
\setlength{\tabcolsep}{4pt}
\resizebox{\textwidth}{!}{%
\begin{tabular}{llccccc}
\toprule
\textbf{Model} & \textbf{Method} & \textbf{Peak (\%)} & \textbf{Peak step} & \textbf{Collapse step} & \textbf{Step-340 (\%)} & \textbf{Retention (\%)} \\
               &                 & {\footnotesize GSM8K\,/\,M500\,/\,MMLU} & {\footnotesize GSM8K\,/\,M500\,/\,MMLU} & {\footnotesize GSM8K\,/\,M500\,/\,MMLU} & {\footnotesize GSM8K\,/\,M500\,/\,MMLU} & {\footnotesize GSM8K\,/\,M500\,/\,MMLU} \\
\midrule
\multirow{4}{*}{Qwen2.5-1.5B}
& GRPO with ground-truth reward                   & 74.9\,/\,59.0\,/\,32.6 & 160\,/\,120\,/\,340  & ---\,/\,---\,/\,---     & 73.2\,/\,56.0\,/\,32.6     & 97.7\,/\,94.9\,/\,100.0 \\
& INTUITOR                            & 72.9\,/\,51.4\,/\,31.0 &  40\,/\, 60\,/\, 40  & 160\,/\,160\,/\,220     &  0.0\,/\, 0.0\,/\, 3.2     &  0.0\,/\, 0.0\,/\,10.4 \\
& Multi-reward (ours)                             & \textbf{73.2}\,/\,53.6\,/\,31.0 &  40\,/\, 40\,/\, 40  & 260\,/\,240\,/\,---     &  1.1\,/\, 2.4\,/\,19.9     &  1.6\,/\, 4.5\,/\,64.2 \\
& Multi-reward + KL-Cov (ours) $\beta_{cov}{=}0.05$   & 73.0\,/\,\textbf{55.0}\,/\,\textbf{31.4} &  60\,/\,140\,/\, 60  & ---\,/\,---\,/\,---     & \textbf{64.9}\,/\,\textbf{50.6}\,/\,\textbf{29.3}     & 88.9\,/\,92.0\,/\,93.3 \\
\midrule
\multirow{4}{*}{Qwen2.5-3B}
& GRPO with ground-truth reward                   & 87.0\,/\,66.0\,/\,41.5 & 260\,/\, 60\,/\,220  & ---\,/\,---\,/\,---     & 85.3\,/\,65.2\,/\,40.7     & 98.0\,/\,98.8\,/\,98.0 \\
& INTUITOR                            & 82.5\,/\,61.2\,/\,39.3 &  40\,/\, 40\,/\, 40  & 280\,/\,280\,/\,---     &  0.0\,/\, 0.8\,/\,28.0     &  0.0\,/\, 1.3\,/\,71.4 \\
& Multi-reward (ours)                             & 83.6\,/\,\textbf{64.6}\,/\,\textbf{40.2} &  60\,/\, 80\,/\, 60  & ---\,/\,---\,/\,---     & 80.4\,/\,47.4\,/\,35.7     & 96.2\,/\,73.4\,/\,88.7 \\
& Multi-reward + KL-Cov (ours) $\beta_{cov}{=}0.05$   & \textbf{84.2}\,/\,64.0\,/\,40.0 &  40\,/\, 40\,/\,260  & ---\,/\,---\,/\,---     & 80.1\,/\,\textbf{62.2}\,/\,\textbf{39.8}     & 95.2\,/\,97.2\,/\,99.5 \\
\bottomrule
\end{tabular}%
}
\end{table}

\paragraph{Three observations from Figure~\ref{fig:trajectory} and Table~\ref{tab:stability}.}

\textbf{(1) Single-signal methods collapse catastrophically.}
INTUITOR peaks at step 40 with $72.9\%$ pass@1 and then declines monotonically: $64.8\%$ (step 80) $\to 44.9\%$ (step 140) $\to 22.4\%$ (step 160) $\to 0\%$ from step 200 onward. By the end of training the policy produces no parseable answers on GSM8K. This matches the theoretical prediction of~\citet{nfl, he2026unsupervised, srt}: an objective that depends on a single reward signal can be maximized by degenerate policies that emit the same high-confidence (but content-free) template regardless of input.

\textbf{(2) Multi-reward without KL-Cov delays collapse on 1.5B but is not consistently stable.}
Plain multi-reward (cluster voting + self-certainty, no KL-Cov) reaches a competitive early peak of $73.2\%$ (step 40) on GSM8K and $53.6\%$ (step 40) on MATH500, and holds this level for considerably longer than either single-signal method: GSM8K stays in the $65$--$73\%$ band through step 200 (and above $56\%$ through step 240), while MATH500 stays above $46\%$ through step 180. Around step 240 it enters a rapid degeneration phase---GSM8K drops from $56.6\%$ at step 240 to $1.1\%$ at step 260---and the policy collapses to $1.1\%$ / $2.4\%$ on GSM8K / MATH500 by step 260, staying there through step 340. The complementary reward structure (cluster voting and self-certainty, with complementary policy gradients per) thus \emph{slows} convergence to the degenerate fixed point, extending the usable training window by ${\approx}60$ steps versus INTUITOR, but does not, by itself, prevent collapse.

\textbf{(3) Multi-reward + KL-Cov is the only unsupervised configuration that remains robust across both scales.}
Multi-reward + KL-Cov ($\beta{=}0.05$) peaks at $73.0\%$ GSM8K (step 60) and $55.0\%$ MATH500 (step 140) on 1.5B, then remains in the $64.9$--$73.0\%$ band on GSM8K and $49.2$--$55.0\%$ band on MATH500 through step 340. On 3B it is even more stable, retaining $95.2\%$ / $97.2\%$ / $99.5\%$ of peak on GSM8K / MATH500 / MMLU-Pro. The trajectories closely track supervised GRPO-GT, supporting the claim that KL-Cov is the ingredient that suppresses the residual gradient concentration on outlier tokens that plain multi-reward alone cannot control.

\section{Component ablation}
\label{app:component_ablation}

Table~\ref{tab:ablation} reports the component-removal ablation summarized in Section~\ref{sec:ablation}. Peaks differ by less than $2$ points across configurations; the diagnostic difference is at step~340, where every reduced configuration collapses on math while the full method retains $\geq 50\%$.

\begin{table}[ht]
\caption{Component ablation on Qwen2.5-1.5B (\textbf{peak\,/\,step-340} pass@1, \%). Peaks are tight; step-340 is the diagnostic. ${}^{\dagger}$Cluster-only was trained to step~280; math accuracy has been flat at the reported value since step~140 (Figure~\ref{fig:single_multi_reward}).}
\label{tab:ablation}
\centering
\begin{tabular}{lcc}
\toprule
\textbf{Configuration} & \textbf{GSM8K} & \textbf{MATH500} \\
                       & {\footnotesize peak / step-340} & {\footnotesize peak / step-340} \\
\midrule
Full model (all components)                    & \textbf{73.0} / \textbf{64.9} & \textbf{55.0} / \textbf{50.6} \\
\midrule
w/o KL-Cov (multi-reward only, ours)                 & 73.2 / 1.1   & 53.6 / 2.4   \\
w/o cluster voting (self-certainty only $\equiv$ INTUITOR)     & 72.9 / 0.0   & 51.4 / 0.0   \\
w/o self-certainty (cluster voting only)       & 75.0 / 0.0${}^{\dagger}$ & 55.6 / 1.4${}^{\dagger}$ \\
\bottomrule
\end{tabular}
\end{table}

\section{Detailed KL-Cov \texorpdfstring{$\beta_{\text{cov}}$}{beta-cov} sweep and KL-Cov on single-signal INTUITOR}
\label{app:klcov_sweep}

This appendix accompanies Figure~\ref{fig:klcov_ablation} (Section~\ref{sec:ablation}) with the per-method numerical breakdown and the complementary INTUITOR$+$KL-Cov ablation referenced in the main text.

\paragraph{Numerical breakdown of the $\beta_{\text{cov}}$ sweep.}
On the multi-reward base, peaks fall in a narrow $72.1$--$73.0\%$ band on GSM8K and $54.8$--$55.0\%$ on MATH500; step-160 accuracy lies within $67.8$--$70.5\%$ on GSM8K and $51.6$--$53.8\%$ on MATH500---${\sim}2.7$ percentage points of step-160 spread across a $200{\times}$ range of $\beta_{\text{cov}}$ (Table~\ref{tab:klcov_sweep}).

\paragraph{KL-Cov on single-signal INTUITOR.}
Does KL-Cov alone---on top of a single-signal reward---suffice to prevent collapse? It does not (Figure~\ref{fig:intuitor_klcov_ablation}). Adding KL-Cov ($\beta_{\text{cov}}{=}0.02$) to vanilla INTUITOR leaves peaks essentially unchanged ($72.6$ vs.\ $72.9$ on GSM8K, $51.2$ vs.\ $51.4$ on MATH500) but doubles step-160 retention ($46.2\%$ / $38.2\%$ vs.\ $22.4\%$ / $22.6\%$). Past step~200, however, the policy still crashes to ${<}1\%$ on both benchmarks: on a single-signal reward the degenerate fixed point is too strong an attractor for KL-Cov alone to escape. The mean-completion-length panel shows the mechanism---KL-Cov delays the collapse into the short template-like regime, but does not prevent the eventual fall.

\begin{figure}[ht]
\centering
\includegraphics[width=0.95\textwidth]{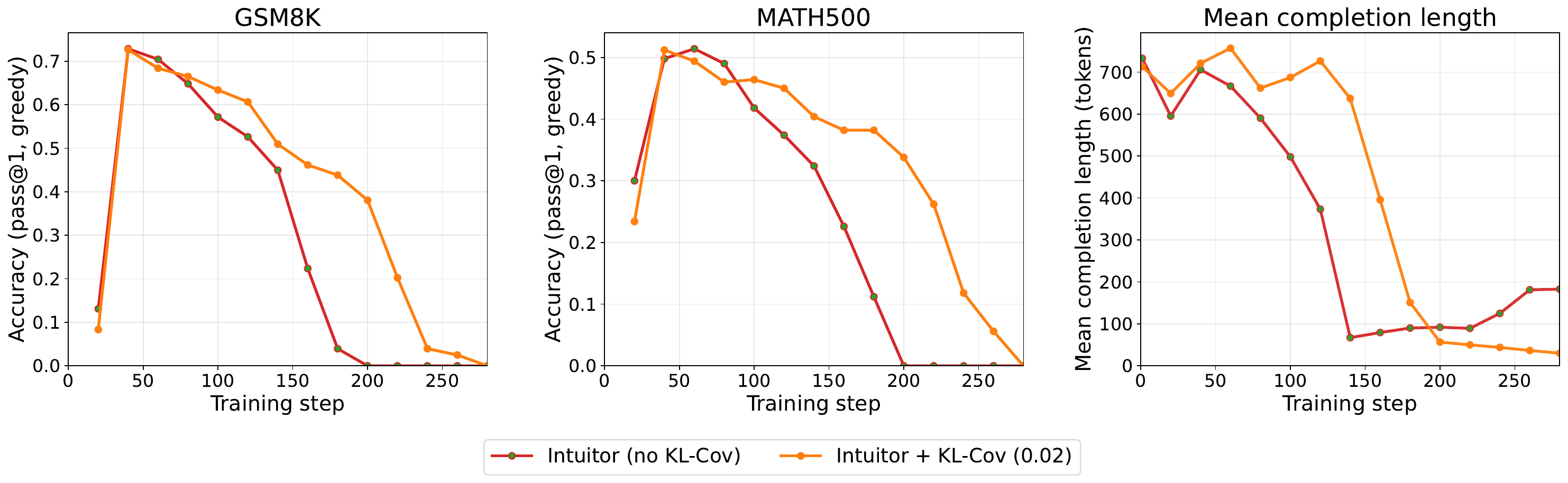}
\caption{INTUITOR + KL-Cov ablation on Qwen2.5-1.5B (steps 0--280). KL-Cov ($\beta_{\text{cov}}{=}0.02$) doubles step-160 accuracy versus vanilla INTUITOR but does not prevent collapse past step~200.}
\label{fig:intuitor_klcov_ablation}
\end{figure}

\begin{table}[ht]
\caption{KL-Cov sweep merged across the INTUITOR (single-signal) and multi-reward base methods on Qwen2.5-1.5B, reporting \textbf{peak\,/\,step-160} pass@1 (\%). The peak captures best-case accuracy; the step-160 column reflects mid-horizon stability under continued training. Within each base reward (top: INTUITOR sweep; bottom: multi-reward sweep), KL-Cov is the only variable. Bold marks the chosen full-method configuration. \emph{Top block}: KL-Cov on single-signal INTUITOR roughly doubles step-160 retention on math benchmarks (e.g., GSM8K $22.4 \to 46.2\%$) but cannot keep pace with the complementary reward structure of multi-reward, and on a longer horizon ($\geq 200$ steps) eventually collapses too. \emph{Bottom block}: KL-Cov is remarkably insensitive to $\beta_{\text{cov}}$---a $200{\times}$ range produces only ${\sim}2.7$ pp of step-160 spread.}
\label{tab:klcov_sweep}
\centering
\small
\begin{tabular}{lccc}
\toprule
\textbf{Configuration} & \textbf{GSM8K} & \textbf{MATH500} & \textbf{MMLU-Pro} \\
                       & {\footnotesize peak / step-160} & {\footnotesize peak / step-160} & {\footnotesize peak / step-160} \\
\midrule
\multicolumn{4}{l}{\textit{Single-signal reward (INTUITOR)}} \\
INTUITOR (no KL-Cov)                            & 72.9 / 22.4 & 51.4 / 22.6 & 31.0 / 24.5 \\
INTUITOR + KL-Cov ($\beta_{\text{cov}}{=}0.02$) & 72.6 / \textbf{46.2} & 51.2 /\textbf{ 38.2} & 30.9 / \textbf{24.9} \\
\midrule
\multicolumn{4}{l}{\textit{Multi-reward (cluster voting + self-certainty + GDPO) --- ours}} \\
Multi-reward (no KL-Cov)                                       & 73.2 / 69.5 & 53.6 / 46.0 & 31.0 / 27.7 \\
Multi-reward + KL-Cov ($\beta_{\text{cov}}{=}0.0005$)          & 72.6 / 67.8 & 54.8 / 52.8 & 31.6 / 30.7 \\
{Multi-reward + KL-Cov ($\beta_{\text{cov}}{=}0.05$)} (chosen) & \textbf{73.0} / 68.3 & \textbf{55.0} / 51.6 & {31.4 / 29.9} \\
Multi-reward + KL-Cov ($\beta_{\text{cov}}{=}0.1$)             & 72.1 / \textbf{70.5} & 54.8 / \textbf{53.8} & \textbf{31.8} / \textbf{30.4} \\
\bottomrule
\end{tabular}
\end{table}

\section{Per-channel GDPO normalization vs.\ na\"{i}ve combination}
\label{app:gdpo_ablation}

This appendix isolates the contribution of per-channel GDPO normalization in the multi-reward advantage of Eq.~\ref{eq:gdpo_norm}--\ref{eq:advantage}. We compare two policy-update variants on Qwen2.5-1.5B with the multi-reward + KL-Cov ($\beta_{\text{cov}}{=}0.05$) objective; everything except the advantage construction is held fixed (same prompts, rollouts, KL-Cov regularizer, optimizer, and seeds).

\paragraph{Variants.}
Let $R^{(\text{cf})}_g$ denote the (cluster$+$format) channel and $R^{(\text{sc})}_g$ the self-certainty channel for rollout $g$ in a $G$-sample group, and let $\hat{z}(\cdot)$ denote per-group $z$-score normalization.
\begin{itemize}
\item \emph{Per-channel GDPO norm (ours).} Each channel is $z$-scored \emph{within its own group} before being combined: $A_g = w_{\text{cf}}\,\hat{z}(R^{(\text{cf})})_g + w_{\text{sc}}\,\hat{z}(R^{(\text{sc})})_g$.
\item \emph{Combine-then-normalize.} Raw weighted rewards are summed at their native scales \emph{first}, and a single $z$-score is applied to the combined signal: $A_g = \hat{z}\!\big(w_{\text{cf}} R^{(\text{cf})} + w_{\text{sc}} R^{(\text{sc})}\big)_g$.
\end{itemize}
Both variants feed the policy update an advantage with std~${\approx}1$, so any difference is attributable to the order of normalization, not to the magnitude of the gradient.

\paragraph{Result (downstream accuracy).}
Per-channel normalization holds its peak through training, while the na\"{i}ve combination peaks early and drifts down (Figure~\ref{fig:gdpo_ablation}). The gap is sharpest on GSM8K---peak $73.0\%$ (per-channel) vs.\ $69.4\%$ (combine-then-normalize), step-280 $67.2\%$ vs.\ $46.5\%$---and consistent on MATH500 and MMLU-Pro. Mechanistically, when the two channels are summed at their native scales before normalization, the higher-variance channel dominates the combined signal: the single $z$-score then implicitly down-weights the lower-variance channel by exactly the ratio of pre-normalization standard deviations. Per-channel $z$-scoring removes this implicit re-weighting and lets $w_{\text{cf}}, w_{\text{sc}}$ control the channel mix as intended.

\begin{figure}[ht]
\centering
\includegraphics[width=\textwidth]{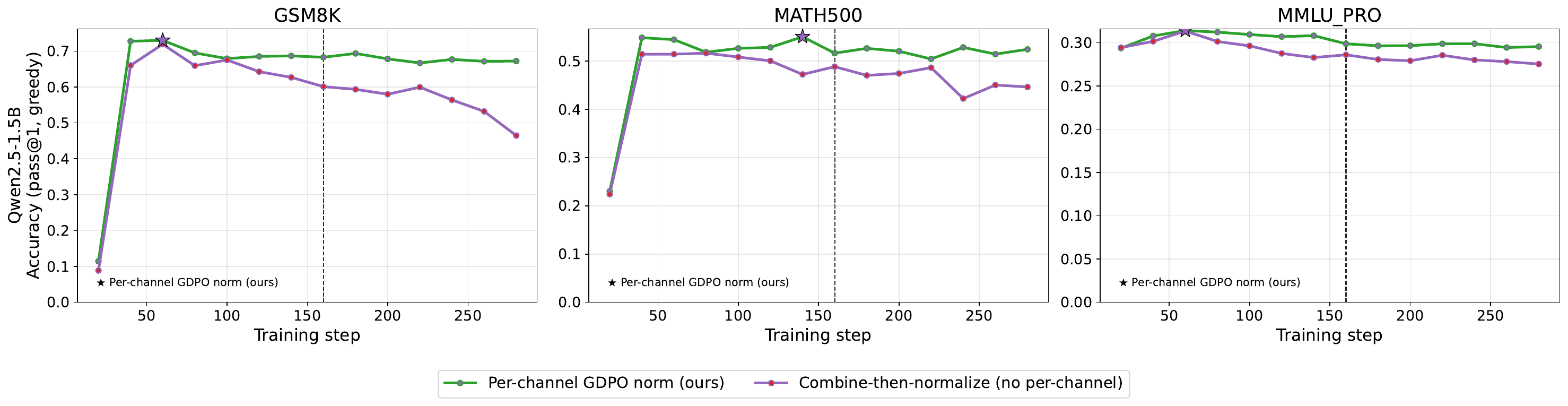}
\caption{GDPO-normalization ablation on Qwen2.5-1.5B with the full multi-reward + KL-Cov ($\beta_{\text{cov}}{=}0.05$) objective. Replacing per-channel $z$-score with a single combine-then-normalize step degrades GSM8K substantially and MATH500 / MMLU-Pro consistently across the entire trajectory; the gap widens past step~160 (vertical line). Stars mark the overall peak per benchmark.}
\label{fig:gdpo_ablation}
\end{figure}
\paragraph{Gradient-space diagnosis.}
To confirm that the trajectory gap in Figure~\ref{fig:gdpo_ablation} is driven by
the variance-weighted channel collapse predicted above, and not by an unrelated
optimization artifact, we instrument training with a lightweight magnitude probe
in per-token logprob space. The choice of logprob space rather than parameter
space can be argued by the chain rule: any loss that depends on the policy only
through its per-token logprobs $\ell$ has parameter gradient
$\nabla_\theta \mathcal{L} = J^\top \nabla_\ell \mathcal{L}$, where the Jacobian
$J = \partial \ell / \partial \theta$ depends only on the model and not on which
reward channel produced $\mathcal{L}$. The same $J^\top$ multiplies both channel
gradients at every step, so a $k{\times}$ gap at logprob space passes through to
an approximately $k{\times}$ gap at parameter space (up to anisotropy of
$J^\top$ in the specific directions
$\nabla_\ell \mathcal{L}_{\mathrm{cf}}^{\mathrm{lin}}$ and
$\nabla_\ell \mathcal{L}_{\mathrm{sc}}^{\mathrm{lin}}$ point, which is shared
across channels and therefore symmetric in the cf/sc ratio). The logprob-space
norm ratio is consequently a faithful proxy for the relative magnitude with
which each channel drives the parameter update, computable in
$\mathcal{O}(B \cdot S)$ without any traversal of the model and trivially safe
under FSDP.

\paragraph{Linearization and per-channel decomposition.}
At every training step we linearize the clipped surrogate around its actual
per-token coefficient $c^{\mathrm{eff}}_{g,t}\in\{\rho_{g,t},\bar{\rho}_{g,t}\}$
---the unclipped ratio for unclipped or KL-Cov-gated tokens, the clipped ratio
otherwise. Because both channels share the same $c^{\mathrm{eff}}_{g,t}$ and the
same completion mask $m_{g,t}$, the surrogate splits cleanly into per-channel
pieces
\begin{equation*}
\mathcal{L}_{\mathrm{cf}}^{\mathrm{lin}}
  = \mathbb{E}_{g,t}\!\left[-c^{\mathrm{eff}}_{g,t}\,\hat{R}_{\mathrm{cf}}(g)\,m_{g,t}\right],
\qquad
\mathcal{L}_{\mathrm{sc}}^{\mathrm{lin}}
  = \mathbb{E}_{g,t}\!\left[-c^{\mathrm{eff}}_{g,t}\,\hat{R}_{\mathrm{sc}}(g)\,m_{g,t}\right],
\end{equation*}
which sum to $\mathcal{L}_{\mathrm{policy}}^{\mathrm{clip}} =
w_{\mathrm{cf}}\,\mathcal{L}_{\mathrm{cf}}^{\mathrm{lin}} +
w_{\mathrm{sc}}\,\mathcal{L}_{\mathrm{sc}}^{\mathrm{lin}}$ exactly the
operating point (the global KL term and the KL-Cov $|\log\rho|$ penalty do not
split by channel and are excluded). We back-propagate each piece only as far as
$\ell$ and report the masked, completion-token L2 norms
$\|\nabla_\ell \mathcal{L}_{\mathrm{cf}}^{\mathrm{lin}}\|_2$ and
$\|\nabla_\ell \mathcal{L}_{\mathrm{sc}}^{\mathrm{lin}}\|_2$ at every step.

\paragraph{Result.}
Figure~\ref{fig:gdpo_gradients} shows the two channels' logprob-space gradient
norms across training under both normalization variants. Under
combine-then-normalize (purple),
$\|\nabla_\ell \mathcal{L}_{\mathrm{sc}}^{\mathrm{lin}}\|$ sits $10\text{--}20{\times}$
above $\|\nabla_\ell \mathcal{L}_{\mathrm{cf}}^{\mathrm{lin}}\|$ throughout
training; equivalently, the cf channel is delivering a parameter update an order
of magnitude weaker than sc at every step. Under per-channel GDPO (green) the
two norms sit on the same order of magnitude, with cf in fact slightly above sc.
This is the variance-weighted channel collapse predicted analytically: when raw
rewards are summed before normalization, the single combined $z$-score
implicitly down-weights the lower-variance channel by exactly the ratio
$\sigma_{\mathrm{cf}}/\sigma_{\mathrm{sc}}$ of pre-normalization standard
deviations, and the gradient the policy actually receives from cf is suppressed
by that same factor. Per-channel $z$-scoring removes this implicit re-weighting
at the source, restoring both channels to the relative influence specified by
$w_{\mathrm{cf}}$ and $w_{\mathrm{sc}}$. The trajectory degradation in
Figure~\ref{fig:gdpo_ablation} is therefore the
predictable downstream consequence of letting one channel's pre-normalization
variance silently dictate the channel mix.

\begin{figure}[ht]
\centering
\includegraphics[width=1\textwidth]{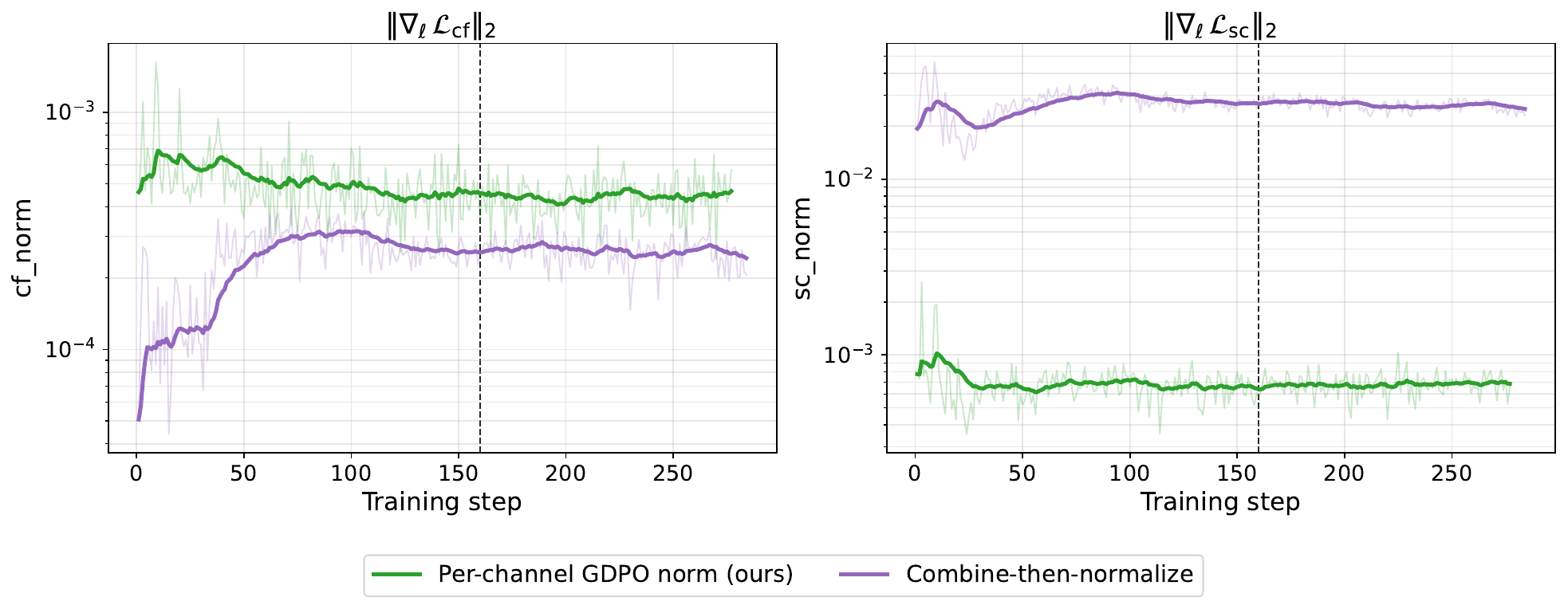}
\caption{Logprob-space gradient norms per reward channel under each
normalization variant. \textbf{Left:} cluster+format channel norm. Under
combine-then-normalize (purple) the cf channel is roughly an order of magnitude
weaker than under per-channel GDPO (green); the channel is being effectively
silenced by the na\"{i}ve combination. \textbf{Right:} self-certainty channel
norm. Combine-then-normalize is $10\text{--}20{\times}$ larger, i.e.\ the
higher-variance channel dominates the policy update by precisely the ratio of
pre-normalization standard deviations. Both panels use a logarithmic $y$-axis.
Vertical dashed line at step~160 matches Figure~\ref{fig:gdpo_ablation}.}
\label{fig:gdpo_gradients}
\end{figure}

\section{Hyperparameters}
\label{app:hyperparams}

Table~\ref{tab:hyperparams} reports the full set of training and evaluation hyperparameters. The same configuration is used across all unsupervised configurations and the GRPO-GT supervised baseline; only the reward composition, $\beta_{\text{cov}}$ (KL-Cov coefficient), and the per-device batch / gradient-accumulation pair (smaller per-device batch on 3B for memory, with grad-accum increased to keep the effective batch fixed) vary across runs. For the $\beta_{\text{cov}}$ sweep (Appendix~\ref{app:klcov_sweep}) we use $\beta_{\text{cov}} \in \{0.0005, 0.05, 0.1\}$ on the multi-reward objective; the INTUITOR$+$KL-Cov ablation uses $\beta_{\text{cov}}{=}0.02$.

\begin{table}[ht]
\caption{Training, optimization, and evaluation hyperparameters used across all configurations. Values are shared between Qwen2.5-1.5B and Qwen2.5-3B unless explicitly listed per-model. The format-reward weight applies only to the multi-reward configurations that include the boxed-format auxiliary; KL-Cov entries apply only when KL-Cov is on.}
\label{tab:hyperparams}
\centering
\small
\begin{tabular}{lcc}
\toprule
\textbf{Hyperparameter} & \textbf{Qwen2.5-1.5B} & \textbf{Qwen2.5-3B} \\
\midrule
\multicolumn{3}{l}{\emph{Optimization}} \\
Base optimizer            & AdamW             & AdamW \\
Learning rate             & $3\times 10^{-6}$ & $3\times 10^{-6}$ \\
LR schedule               & cosine            & cosine \\
Warmup ratio              & 0.1               & 0.1 \\
Max gradient norm         & 1.0               & 1.0 \\
\midrule
\multicolumn{3}{l}{\emph{GRPO objective}} \\
Rollouts per prompt $G$            & 7     & 7 \\
Sampling temperature (training)    & 1.0   & 1.0 \\
Max completion length              & 3072  & 3072 \\
PPO clip $\epsilon$                & 0.2   & 0.2 \\
Reference-KL coefficient $\gamma$  & 0.005 & 0.005 \\
\midrule
\multicolumn{3}{l}{\emph{Multi-reward and KL-Cov (ours)}} \\
Cluster-reward weight $\alpha$               & 0.5  & 0.5 \\
Self-certainty weight $\beta$                & 0.5  & 0.5 \\
Format-reward weight                         & 0.1  & 0.1 \\
KL-Cov coefficient $\beta_{\text{cov}}$      & 0.05 & 0.05 \\
KL-Cov top-$k$ fraction                      & 0.02 & 0.02 \\
\midrule
\multicolumn{3}{l}{\emph{Batch and schedule}} \\
Per-device train batch          & 21      & 14 \\
Gradient accumulation steps     & 14      & 21 \\
Trainer GPUs (ZeRO-3)           & 3       & 3 \\
Effective batch (completions/step) & 882 & 882 \\
Prompts per gradient step       & 126     & 126 \\
Trained / evaluated horizon     & 340 steps & 340 steps \\
Checkpoint interval             & 20 steps & 20 steps \\
\midrule
\multicolumn{3}{l}{\emph{Hardware (per node)}} \\
GPUs                            & 4$\times$ NVIDIA GH200 96GB & 4$\times$ NVIDIA GH200 96GB \\
Rollout server                  & GPU 0 (vLLM)                 & GPU 0 (vLLM) \\
Trainer                         & GPUs 1--3 (DeepSpeed ZeRO-3) & GPUs 1--3 (DeepSpeed ZeRO-3) \\
\midrule
\multicolumn{3}{l}{\emph{Evaluation}} \\
Decoding                        & greedy ($T{=}0$, $n{=}1$) & greedy ($T{=}0$, $n{=}1$) \\
Math scorer                     & \texttt{math\_verify}     & \texttt{math\_verify} \\
MMLU-Pro prompt                 & TIGER-Lab 5-shot CoT      & TIGER-Lab 5-shot CoT \\
LCB v6 scorer                   & \texttt{lighteval} (all-tests-pass) & \texttt{lighteval} (all-tests-pass) \\
CRUXEval-O prompt               & ZeroEval \texttt{OEQA}    & ZeroEval \texttt{OEQA} \\
\bottomrule
\end{tabular}
\end{table}

\section{Qualitative completions: INTUITOR vs.\ multi-reward on GSM8K (step 240)}
\label{app:qualitative}

To make the late-horizon collapse concrete, we sample five GSM8K problems at the step-240 checkpoint of Qwen2.5-1.5B, where INTUITOR has fully collapsed (GSM8K pass@1 ${=}0.0\%$, mean completion length 282 chars) while multi-reward is still functional (pass@1 ${=}56.6\%$, mean length 1{,}260 chars). At this step there are $747$ problems on which INTUITOR is wrong and multi-reward is correct ($56.6\%$ of the test set); the five problems below have the shortest INTUITOR completions among these. Both INTUITOR and multi-reward completions are reproduced \emph{verbatim} from the saved evaluation logs: Markdown bold/headers are rendered as bold paragraph markers, and the model's LaTeX-style math (\verb|\(...\)|, \verb|\[...\]|, \verb|\boxed{...}|) is typeset as written. The same self-repeating header---``\textit{Let's break it down step-by-step \dots\ Step~1: Determine X. Let's proceed step-by-step. Step~1: Determine X.}''---recurs across all five problems (and across most of the dataset); INTUITOR has converged onto a deterministic short template that never emits a final answer.

\subsection{Problem \texttt{gsm8k-499} (gold: 10)}
\label{app:gsm8k_499}

\paragraph{Question.}
Mark is trying to choose between two venues for a surprise party for his wife. The first venue charges a flat fee of \$200, regardless of how many guests attend. While the second charges, \$25 per person who attends. However, the first venue does not include food, which Mark estimates will cost \$5 for each person who attends. At the second venue, food for each guest is already included in the price. How many guests are necessary for the two venues to be equal in cost?

\begin{longtable}{|p{0.22\textwidth}|p{0.70\textwidth}|} 
\toprule
\textbf{INTUITOR (\textcolor{red}{wrong})} & \textbf{Multi-Reward (correct, $\boxed{10}$)} \\
\midrule
\endfirsthead
% \toprule
% % \textbf{INTUITOR (continued)} & \textbf{Multi-Reward (continued)} \\
% \midrule
\endhead
\bottomrule
\endfoot

\small\itshape
To solve this problem, we need to follow a systematic approach. Let's break it down step-by-step:\par\smallskip
\textbf{Step 1: Define variables for the number of guests.}\par\smallskip
Let $x$ be the number of guests.
&
\small\itshape
To determine the number of guests necessary for the two venues to be equal in cost, we need to set up equations for the total cost of each venue and then solve for the number of guests.\par\smallskip
Let's denote the number of guests by $x$.\par\smallskip
The cost of the first venue is given by:
$$\text{Cost}_1 = 200.$$
The cost of the second venue is given by:
$$\text{Cost}_2 = 25x + 5x = 30x.$$
We need to find the value of $x$ for which the costs are equal. Therefore, we set up the equation:
$$200 = 30x.$$
To solve for $x$, we divide both sides of the equation by 30:
$$x = \frac{200}{30}.$$
Simplifying the fraction, we get:
$$x = \frac{20}{3}.$$
Therefore, the number of guests necessary for the two venues to be equal in cost is:
$$\boxed{10}.$$ \\

\end{longtable}

\subsection{Problem \texttt{gsm8k-644} (gold: 29)}
\label{app:gsm8k_644}

\paragraph{Question.}
Steve is 60 years old. His wife is 4 years older than him. Their son is currently half as old as his mom and their son's wife is 3 years younger than her husband. How old is Steve's son's wife?

\begin{longtable}{|p{0.22\textwidth}|p{0.70\textwidth}|} 
\toprule
\textbf{INTUITOR (\textcolor{red}{wrong})} & \textbf{Multi-Reward (correct, $\boxed{29}$)} \\
\midrule
\endfirsthead
% \toprule
% \textbf{INTUITOR (continued)} & \textbf{Multi-Reward (continued)} \\
% \midrule
\endhead
\bottomrule
\endfoot

\small\itshape
To solve this problem, we need to follow a systematic approach. Let's break it down step-by-step:\par\smallskip
\textbf{Step 1: Determine Steve's age.}\par\smallskip
Let's proceed step-by-step:\par\smallskip
\textbf{Step 1: Determine Steve's age.}
&
\small\itshape
To determine the age of Steve's son's wife, we need to follow these steps:
\begin{enumerate}\itemsep0pt
\item Identify the current age of Steve.
\item Calculate the current age of Steve's wife.
\item Determine the current age of Steve's son.
\item Calculate the current age of Steve's son's wife.
\end{enumerate}
Let's go through each step in detail.\par\smallskip
\textbf{Step 1: Identify the current age of Steve.} Steve is currently 60 years old.\par\smallskip
\textbf{Step 2: Calculate the current age of Steve's wife.} Steve's wife is 4 years older than Steve. Therefore, her age is:
$$\text{Age of Steve's wife} = \text{Age of Steve} + 4 = 60 + 4 = 64 \text{ years old}.$$
\textbf{Step 3: Determine the current age of Steve's son.} Steve's son is currently half as old as his mom. Therefore, his age is:
$$\text{Age of Steve's son} = \frac{\text{Age of Steve's wife}}{2} = \frac{64}{2} = 32 \text{ years old}.$$
\textbf{Step 4: Calculate the current age of Steve's son's wife.} Steve's son's wife is 3 years younger than her husband. Therefore, her age is:
$$\text{Age of Steve's son's wife} = \text{Age of Steve's son} - 3 = 32 - 3 = 29 \text{ years old}.$$
Thus, the age of Steve's son's wife is $\boxed{29}$. \\

\end{longtable}

\subsection{Problem \texttt{gsm8k-895} (gold: 74)}
\label{app:gsm8k_895}

\paragraph{Question.}
Martin's weight is 55~kg. Carl's weight is 16~kg more than Martin's weight. Christian's weight is 8~kg more than Carl's weight. Harry is 5~kg less than Christian's weight. What is the weight of Harry, in kg?

\begin{longtable}{|p{0.22\textwidth}|p{0.70\textwidth}|} 
\toprule
\textbf{INTUITOR (\textcolor{red}{wrong})} & \textbf{Multi-Reward (correct, $\boxed{74}$)} \\
\midrule
\endfirsthead
% \toprule
% \textbf{INTUITOR (continued)} & \textbf{Multi-Reward (continued)} \\
% \midrule
\endhead
\bottomrule
\endfoot

\small\itshape
To solve this problem, we need to follow a systematic approach. Let's break it down step-by-step:\par\smallskip
\textbf{Step 1: Determine Carl's weight.}\par\smallskip
Let's proceed step-by-step:\par\smallskip
\textbf{Step 1: Determine Carl's weight.}
&
\small\itshape
To determine the weight of Harry, we need to follow a step-by-step approach. Let's break it down.
\begin{enumerate}\itemsep0pt
\item Identify Martin's weight.
\item Calculate Carl's weight.
\item Calculate Christian's weight.
\item Determine Harry's weight.
\end{enumerate}
Let's go through each step in detail.\par\smallskip
\textbf{Step 1: Identify Martin's weight.} Martin's weight is given as 55~kg.\par\smallskip
\textbf{Step 2: Calculate Carl's weight.} Carl's weight is 16~kg more than Martin's weight. So, we can calculate Carl's weight as follows:
$$\text{Carl's weight} = \text{Martin's weight} + 16 = 55 + 16 = 71 \text{ kg}.$$
\textbf{Step 3: Calculate Christian's weight.} Christian's weight is 8~kg more than Carl's weight. So, we can calculate Christian's weight as follows:
$$\text{Christian's weight} = \text{Carl's weight} + 8 = 71 + 8 = 79 \text{ kg}.$$
\textbf{Step 4: Determine Harry's weight.} Harry's weight is 5~kg less than Christian's weight. So, we can calculate Harry's weight as follows:
$$\text{Harry's weight} = \text{Christian's weight} - 5 = 79 - 5 = 74 \text{ kg}.$$
Therefore, the weight of Harry is $\boxed{74}$~kg. \\

\end{longtable}

\subsection{Problem \texttt{gsm8k-561} (gold: 180)}
\label{app:gsm8k_561}

\paragraph{Question.}
If Mark weighs 150 pounds and Susan weighs 20 pounds less than Mark. And their friend Bob weighs twice as much as Susan. What is the average weight of the 3 friends?

\begin{longtable}{|p{0.22\textwidth}|p{0.70\textwidth}|} 
\toprule
\textbf{INTUITOR (\textcolor{red}{wrong})} & \textbf{Multi-Reward ({correct}, $\boxed{180}$)} \\
\midrule
\endfirsthead
% \toprule
% \textbf{INTUITOR (continued)} & \textbf{Multi-Reward (continued)} \\
% \midrule
\endhead
\bottomrule
\endfoot

\small\itshape
To solve this problem, we need to follow a systematic approach. Let's break it down step-by-step:\par\smallskip
\textbf{Step 1: Determine Susan's weight.}\par\smallskip
Let's proceed step-by-step:\par\smallskip
\textbf{Step 1: Determine Susan's weight.}
&
\small\itshape
To determine the average weight of Mark, Susan, and Bob, we need to follow these steps:
\begin{enumerate}\itemsep0pt
\item Calculate Susan's weight.
\item Calculate Bob's weight.
\item Find the total weight of all three friends.
\item Divide the total weight by the number of friends to find the average weight.
\end{enumerate}
Let's go through each step in detail.\par\smallskip
\textbf{Step 1: Calculate Susan's weight.} Mark weighs 150~pounds. Susan weighs 20~pounds less than Mark. Therefore, we can calculate Susan's weight as follows:
$$\text{Susan's weight} = \text{Mark's weight} - 20 = 150 - 20 = 130 \text{ pounds}.$$
\textbf{Step 2: Calculate Bob's weight.} Bob weighs twice as much as Susan. Therefore, we can calculate Bob's weight as follows:
$$\text{Bob's weight} = 2 \times \text{Susan's weight} = 2 \times 130 = 260 \text{ pounds}.$$
\textbf{Step 3: Find the total weight of all three friends.} The total weight is the sum of the weights of Mark, Susan, and Bob:
$$\text{Total weight} = 150 + 130 + 260.$$
Let's calculate the sum:
$$150 + 130 = 280, \qquad 280 + 260 = 540.$$
So, the total weight of all three friends is 540~pounds.\par\smallskip
\textbf{Step 4: Divide the total weight by the number of friends to find the average weight.} There are 3 friends, so we divide the total weight by 3:
$$\text{Average weight} = \frac{540}{3} = 180.$$
Therefore, the average weight of the three friends is $\boxed{180}$~pounds. \\

\end{longtable}

\subsection{Problem \texttt{gsm8k-967} (gold: 30)}
\label{app:gsm8k_967}

\paragraph{Question.}
Nani is 8 years old. His brother is twice his age. Nani's sister is 25\% younger than him. What is the total age of all three of the family members?

\begin{longtable}{|p{0.22\textwidth}|p{0.70\textwidth}|} 
% {p{0.22\textwidth} p{0.70\textwidth}}
\toprule
\textbf{INTUITOR \textcolor{red}{(wrong)}} & \textbf{Multi-Reward (correct, $\boxed{30}$)} \\
\midrule
\endfirsthead
% \toprule
% \textbf{INTUITOR (continued)} & \textbf{Multi-Reward (continued)} \\
% \midrule
\endhead
\bottomrule
\endfoot

\small\itshape
To solve this problem, we need to follow a systematic approach. Let's break it down step-by-step:\par\smallskip
\textbf{Step 1: Determine Nani's brother's age.}\par\smallskip
Let's proceed step-by-step:\par\smallskip
\textbf{Step 1: Determine Nani's brother's age.}
&
\small\itshape
To determine the total age of all three family members, we need to follow these steps:
\begin{enumerate}\itemsep0pt
\item Identify the age of Nani.
\item Calculate the age of Nani's brother.
\item Calculate the age of Nani's sister.
\item Sum the ages of Nani, his brother, and his sister.
\end{enumerate}
Let's go through each step in detail.\par\smallskip
\textbf{Step 1: Identify the age of Nani.} Nani is 8 years old.\par\smallskip
\textbf{Step 2: Calculate the age of Nani's brother.} Nani's brother is twice Nani's age. So, we calculate:
$$\text{Age of Nani's brother} = 2 \times \text{Age of Nani} = 2 \times 8 = 16 \text{ years}.$$
\textbf{Step 3: Calculate the age of Nani's sister.} Nani's sister is 25\% younger than Nani. This means her age is 75\% of Nani's age. So, we calculate:
$$\text{Age of Nani's sister} = 0.75 \times \text{Age of Nani} = 0.75 \times 8 = 6 \text{ years}.$$
\textbf{Step 4: Sum the ages of Nani, his brother, and his sister.} Now, we add the ages together:
$$\text{Total age} = 8 + 16 + 6 = 30.$$
Therefore, the total age of all three family members is $\boxed{30}$. \\

\end{longtable}

\paragraph{What the table shows.} INTUITOR's output is the same self-repeating template across every row: a 4--6-line preamble that announces a "step-by-step" plan and then loops "Step 1: Determine X. Let's proceed step-by-step. Step 1: Determine X." until truncation, never producing a \texttt{\textbackslash boxed\{\}} answer. The token-level entropy of this template is in $[0.03, 0.17]$ nats (Figure~\ref{fig:single_multi_reward}), which is exactly what self-certainty rewards: maximally peaked next-token distributions over a fixed string. Multi-reward trained on the same prompts, the same base model, and the same step count derives the correct answer in every case --- because the cluster channel pays a non-zero reward only on consensus over the \emph{final} answer (Equation~\ref{eq:cluster_reward}), the optimizer cannot collect cluster reward by emitting a confident-but-empty template. This is the per-prompt counterpart of the complementary-failure-modes claim of Section~\ref{sec:ablation}.

\section{Example completions on code-reasoning benchmarks}
\label{app:code_examples}

We complement the GSM8K analysis of Appendix~\ref{app:qualitative} with one INTUITOR-vs-multi-reward case per code benchmark from the step-160 checkpoint of Qwen2.5-1.5B (the same checkpoint scored in Table~\ref{tab:main_results_all}). Both models are trained only on math, so both code benchmarks are out-of-distribution probes for whether the reasoning structure transfers. Both INTUITOR and multi-reward completions are reproduced \emph{verbatim} from the saved evaluation logs (Markdown markers rendered as italics; LaTeX-style math kept as in the model output; only minor formatting such as code-fence rewrapping for column fit).

\subsection{Example from CRUXEval-O (\texttt{crux-test-\#675})}
\label{app:cruxeval_example}

\paragraph{Question.}
You are given a Python function and an assertion containing an input to the function. Complete the assertion with a literal (no unsimplified expressions, no function calls) containing the output when executing the provided code on the given input, even if the function is incorrect or incomplete.

\begin{verbatim}
[PYTHON]
def f(nums, sort_count):
    nums.sort()
    return nums[:sort_count]
assert f([1, 2, 2, 3, 4, 5], 1) == ??
[/PYTHON]
\end{verbatim}

\paragraph{Instruction.}
Please answer this question by first reasoning and then providing your answer. Present your reasoning and solution in a JSON object with \texttt{"reasoning"} and \texttt{"answer"} fields, e.g., \texttt{"answer": "42"}. \emph{Gold answer:} \texttt{[1]}.

\paragraph{INTUITOR Result (incorrect, $\boxed{1}$).}
\begin{quote}\small\itshape
To determine the output of the function \texttt{f([1, 2, 2, 3, 4, 5], 1)}, we need to follow these steps:

\begin{enumerate}\itemsep0pt
\item Sort the list \texttt{[1, 2, 2, 3, 4, 5]}.
\item Return the first element of the sorted list.
\end{enumerate}

Let's proceed step by step.

\textbf{Step 1: Sort the list.} The given list is \texttt{[1, 2, 2, 3, 4, 5]}. When we sort this list, we get
$$[1, 2, 2, 3, 4, 5].$$

\textbf{Step 2: Return the first element.} The first element of the sorted list is \texttt{1}.

Therefore, the output of the function \texttt{f([1, 2, 2, 3, 4, 5], 1)} is \texttt{1}.

So, the final answer is:
$$\boxed{1}$$
\end{quote}

\paragraph{Multi-Reward Result (correct, $\boxed{[1]}$).}
\begin{quote}\small\itshape
To solve this problem, we need to follow these steps:

\begin{enumerate}\itemsep0pt
\item \textbf{Understand the function \texttt{f}.}
\item \textbf{Evaluate the function with the given input.}
\item \textbf{Provide the output of the function.}
\end{enumerate}

\textbf{Step-by-Step Solution.}

\textbf{Step 1: Understand the function \texttt{f}.} The function \texttt{f} takes two parameters: \texttt{nums} and \texttt{sort\_count}. It sorts the list \texttt{nums} in ascending order and then returns the first \texttt{sort\_count} elements of the sorted list.

\textbf{Step 2: Evaluate the function with the given input.} The input to the function is \texttt{f([1, 2, 2, 3, 4, 5], 1)}. Let's break it down:
\begin{itemize}\itemsep0pt
\item The list \texttt{nums} is \texttt{[1, 2, 2, 3, 4, 5]}.
\item The \texttt{sort\_count} is \texttt{1}.
\end{itemize}
First, we sort the list \texttt{nums} in ascending order:
$$\text{sorted\_nums} = [1, 2, 2, 3, 4, 5].$$
Next, we return the first \texttt{sort\_count} elements of the sorted list. Since \texttt{sort\_count} is \texttt{1}, we return the first element of the sorted list:
$$\texttt{f}([1, 2, 2, 3, 4, 5], 1) = [1].$$

\textbf{Step 3: Provide the output of the function.} The output of the function \texttt{f([1, 2, 2, 3, 4, 5], 1)} is \texttt{[1]}.

Therefore, the assertion should be:
$$\boxed{[1]}$$
\end{quote}

\paragraph{What this isolates.} INTUITOR misreads the slice \texttt{nums[:sort\_count]} as "the first element," dropping the list-vs-scalar distinction; the boxed token is a confident integer rather than a list. Multi-reward attends to the slice operator and emits the correct list literal. The two completions are otherwise stylistically similar---this is a one-token semantic error driven by the policy converging on the high-confidence (but wrong) "first element" reading.

\subsection{Example from LiveCodeBench v6 (\texttt{abc344\_a}, AtCoder)}
\label{app:lcb_example}

\paragraph{Question.}
You will be given a question (problem specification) and will generate a correct Python program that matches the specification and passes all tests.

\medskip
\noindent
\emph{Problem.} You are given a string $S$ of lowercase English letters and the character \texttt{|}; $S$ contains exactly two \texttt{|}'s. Remove the characters between the two \texttt{|}'s, including the \texttt{|}'s themselves, and print the resulting string.

\begin{verbatim}
Input
  S
Constraints
  - S has length 2-100, lowercase letters and |.
  - S contains exactly two |'s.

Sample Input 1   atcoder|beginner|contest
Sample Output 1  atcodercontest

Sample Input 2   |spoiler|
Sample Output 2  (empty string)

Sample Input 3   ||xyz
Sample Output 3  xyz
\end{verbatim}

\paragraph{Instruction.}
Read the input from standard input, solve the problem, and write the answer to standard output. Enclose your code within delimiters as follows: \texttt{```python}\,\dots\,\texttt{```}. Scoring: pass@1, all public and private tests must pass.

\paragraph{INTUITOR Result (incorrect; only the \emph{prefix} is kept).}
\begin{quote}\small\itshape
To solve the problem, we need to follow these steps:

\begin{enumerate}\itemsep0pt
\item Read the input string.
\item Remove the characters between the two \texttt{|}'s.
\item Print the resulting string.
\end{enumerate}

Let's proceed step by step.

\textbf{Step 1: Read the input string.} We will use the \texttt{input()} function to read the input string.

\textbf{Step 2: Remove the characters between the two \texttt{|}'s.} We will use string slicing to remove the characters between the two \texttt{|}'s.

\textbf{Step 3: Print the resulting string.} We will use the \texttt{print()} function to output the result.

Here is the Python code to achieve this:
\end{quote}

\begin{verbatim}
# Step 1: Read the input string
input_string = input()

# Step 2: Remove the characters between the two |s
result = input_string.split('|')[0]

# Step 3: Print the resulting string
print(result)
\end{verbatim}

\begin{quote}\small\itshape
Let's run this code to see the output.
\end{quote}

\noindent INTUITOR's solution splits on \texttt{|} and keeps the first piece only. On the sample inputs it returns \texttt{atcoder} (vs.\ \texttt{atcodercontest}), the empty string (correct, by accident), and the empty string (vs.\ \texttt{xyz}). Two of three public tests fail; the submission is scored 0.

\paragraph{Multi-Reward Result (correct).}
\begin{quote}\small\itshape
To solve the problem, we need to follow these steps:

\begin{enumerate}\itemsep0pt
\item Read the input string \texttt{S}.
\item Find the positions of the two \texttt{|} characters.
\item Remove the characters between these two \texttt{|} characters.
\item Print the resulting string.
\end{enumerate}

Here is the Python code that implements these steps:
\end{quote}

\begin{verbatim}
# Read the input string
S = input()

# Find the positions of the two '|'
left_index = S.find('|')
right_index = S.rfind('|')

# Remove the characters between the two '|'
result = S[:left_index] + S[right_index + 1:]

# Print the resulting string
print(result)
\end{verbatim}

\begin{quote}\small\itshape
Let's break down the code:

\begin{enumerate}\itemsep0pt
\item \textbf{Read the input string}: The \texttt{input()} function reads the input from the standard input.
\item \textbf{Find the positions of the two \texttt{|} characters}: The \texttt{find()} method is used to find the first occurrence of the character \texttt{|} in the string \texttt{S}, and the \texttt{rfind()} method is used to find the last occurrence of the character \texttt{|} in the string \texttt{S}.
\item \textbf{Remove the characters between the two \texttt{|} characters}: The string slicing \texttt{S[:left\_index] + S[right\_index + 1:]} is used to remove the characters between the two \texttt{|} characters.
\item \textbf{Print the resulting string}: The \texttt{print()} function is used to print the resulting string.
\end{enumerate}

When you run this code with the provided sample inputs, it will correctly remove the characters between the two \texttt{|} characters and print the resulting string.
\end{quote}

\noindent This passes all three public tests and the private tests (pass@1 = 1.0).

\paragraph{What this isolates.} The semantic gap is precisely the operative phrase "between the \emph{two} \texttt{|}'s." INTUITOR's proposed plan announces this requirement but its emitted code only consumes the prefix before the first \texttt{|},a high-confidence but logically truncated implementation, the same pattern as the GSM8K examples in Appendix~\ref{app:qualitative}. Multi-reward keeps both delimiter positions and emits a slice that satisfies all three sample I/O cases. Because all rollouts must agree on the final answer for the cluster channel to pay reward, the optimizer cannot collect cluster reward via the truncated-plan template that INTUITOR collapses to.

\newpage
\section*{NeurIPS Paper Checklist}

\begin{enumerate}

\item {\bf Claims}
    \item[] Question: Do the main claims made in the abstract and introduction accurately reflect the paper's contributions and scope?
    \item[] Answer: \answerYes{}
    \item[] Justification: The abstract and Section~\ref{sec:introduction} state three contributions: (i)~a multi-reward RLIF framework combining cluster voting and self-certainty with GDPO normalization (Sections~\ref{sec:cluster_voting}--\ref{sec:gdpo}); (ii)~KL-Cov regularization on high-covariance tokens (Section~\ref{sec:klcov}); (iii)~empirical evaluation on five math and two code-reasoning benchmarks and no model collapse in extended training. Each is supported by experiments: Table~\ref{tab:main_results_all} (performance at step 160), Figure~\ref{fig:trajectory} and Table~\ref{tab:scaling} (collapse resistance), Figures~\ref{fig:single_multi_reward} and~\ref{fig:klcov_ablation} (ablations).

\item {\bf Limitations}
    \item[] Question: Does the paper discuss the limitations of the work performed by the authors?
    \item[] Answer: \answerYes{}
    \item[] Justification: Section~\ref{sec:discussion} (\textit{Limitations}) discusses the limitations, including: (i)~the framework adds hyperparameters $\alpha,\beta,\beta_{\text{cov}}$ beyond standard GRPO, with only $\beta_{\text{cov}}$ shown to be insensitive to a $200{\times}$ sweep (Figure~\ref{fig:klcov_ablation}) and $\alpha/\beta$ fixed at the symmetric default $0.5$; (ii)~the cluster-voting reward assumes a single canonical correct answer recoverable by string equivalence, which limits applicability to open-ended generation; (iii)~the method depends on a sufficiently capable base model, since the policy must initially produce some correct rollouts for cluster voting to provide signal.

\item {\bf Theory assumptions and proofs}
    \item[] Question: For each theoretical result, does the paper provide the full set of assumptions and a complete (and correct) proof?
    \item[] Answer: \answerNA{}{}
    \item[] Justification: All the theories used in this paper are cited with proper reference.
    
    % \textcolor{blue}{Theorem~\ref{lem:structural}} (Section~\ref{sec:self_certainty}) states the structurally complementary gradient result with all symbols defined in Section~\ref{sec:preliminaries}.

    \item {\bf Experimental result reproducibility}
    \item[] Question: Does the paper fully disclose all the information needed to reproduce the main experimental results of the paper to the extent that it affects the main claims and/or conclusions of the paper (regardless of whether the code and data are provided or not)?
    \item[] Answer: \answerYes{}
    \item[] Justification: Section~\ref{sec:setup} specifies the base models (Qwen2.5-1.5B-Base, Qwen2.5-3B-Base, Qwen2.5-7B-Base, Qwen3-1.7B-Base), training dataset (\texttt{MATH-lighteval}), evaluation benchmarks, optimizer (GRPO with AdamW), schedule, and key hyperparameters. Appendix~\ref{app:hyperparams} (Table~\ref{tab:hyperparams}) provides the full set of training, optimization, batch, hardware, and evaluation hyperparameters used across all configurations. The covariance estimator and KL-Cov loss are specified by Equations~\ref{eq:covariance}--\ref{eq:total_loss}. Greedy decoding ($T{=}0,n{=}1$), \texttt{math\_verify} scoring, and the lighteval LCB / ZeroEval CRUXEval-O pipelines are documented in the Evaluation paragraph of Section~\ref{sec:setup}.

\item {\bf Open access to data and code}
    \item[] Question: Does the paper provide open access to the data and code, with sufficient instructions to faithfully reproduce the main experimental results, as described in supplemental material?
    \item[] Answer: \answerYes{}
    \item[] Justification: The full training and evaluation code is publicly released under Apache-2.0 at \url{https://github.com/machine-intelligence-lab-wvu/multi-reward}, together with the configuration file for every run reported in the paper and a table mapping each result to its config. All the training and evaluation data are publicly available HuggingFace datasets (\texttt{MATH-lighteval}, \texttt{openai/gsm8k}, \texttt{HuggingFaceH4/MATH-500}, \texttt{HuggingFaceH4/aime\_2024}, \texttt{yentinglin/aime\_2025}, \texttt{TIGER-Lab/MMLU-Pro}, \texttt{lighteval/code\_generation\_lite}, \texttt{flydust/zero-eval}).

\item {\bf Experimental setting/details}
    \item[] Question: Does the paper specify all the training and test details (e.g., data splits, hyperparameters, how they were chosen, type of optimizer) necessary to understand the results?
    \item[] Answer: \answerYes{}
    \item[] Justification: Section~\ref{sec:setup} reports the optimizer (GRPO with AdamW, learning rate $3{\times}10^{-6}$, cosine schedule), rollout configuration ($G{=}7$, max completion length 3072, training temperature 1.0), KL coefficients ($\gamma{=}0.005$ following INTUITOR, $\beta_{\text{cov}}{=}0.05$, top-$k{=}0.02$), reward weights ($\alpha{=}\beta{=}0.5$), batch sizing (882 completions, i.e.\ 126 prompts $\times$ 7 rollouts, per step), and horizon (340 steps, checkpoints every 20 steps). Appendix~\ref{app:hyperparams} (Table~\ref{tab:hyperparams}) lists the complete set of values used per model size, including the per-device batch / gradient-accumulation pair, ZeRO-3 configuration, vLLM rollout server, and evaluation pipeline (greedy $T{=}0,n{=}1$, \texttt{math\_verify}, MMLU-Pro 5-shot CoT, lighteval LCB pipeline, ZeroEval \texttt{OEQA} CRUXEval-O template). Test splits are the standard public test splits of each benchmark; no validation split is used.

\item {\bf Experiment statistical significance}
    \item[] Question: Does the paper report error bars suitably and correctly defined or other appropriate information about the statistical significance of the experiments?
    \item[] Answer: \answerYes{}
    \item[] Justification: {Where the comparison is between two trained arms on the same evaluation set, we report interval estimates and significance tests: Wilson $95\%$ confidence intervals, and an exact McNemar test on paired per-item outcomes, in Appendices~\ref{app:qwen3} and~\ref{app:grpo_klcov} (e.g.\ $p{=}0.030$ for the step-$200$ GSM8K advantage of the KL-Cov arm). We do \emph{not} report error bars over training seeds: each trajectory is a single multi-day run, and replicating every configuration across seeds exceeds our compute budget. This is stated explicitly in the Limitations, and we mitigate it by evaluating every $20$ steps across the full horizon, so the collapse-resistance claim rests on a dense trajectory rather than a single end-point measurement.}
    % Each pass@1 reported in Tables~\ref{tab:main_results}--\ref{tab:code_eval} is therefore a single point estimate; we do not run multiple training seeds because each multi-reward run requires ${\sim}12$--$24$~hours on 4$\times$GH200 GPUs, and the budget would not extend to multi-seed averages across the eight 1.5B configurations and five 3B configurations evaluated. We mitigate this by reporting trajectories evaluated at every 20 training steps (Figure~\ref{fig:trajectory}, Table~\ref{tab:stability}); the rank ordering and collapse-resistance pattern that supports our main claim is consistent across the full 340-step horizon and across both model scales. For AIME 2024/2025 we explicitly note that one problem corresponds to ${\approx}3.3$~pp at the 30-problem benchmark size and report those numbers as far-OOD probes only (Appendix~\ref{app:aime}).
    

\item {\bf Experiments compute resources}
    \item[] Question: For each experiment, does the paper provide sufficient information on the computer resources (type of compute workers, memory, time of execution) needed to reproduce the experiments?
    \item[] Answer: \answerYes{}
    \item[] Justification: Section~\ref{sec:setup} states that training is conducted on a single node with 4 NVIDIA GH200 96GB GPUs. Appendix~\ref{app:hyperparams} (Table~\ref{tab:hyperparams}) breaks down the per-node layout: GPU~0 hosts the vLLM rollout server, GPUs~1--3 run the trainer under DeepSpeed ZeRO-3. Each training run covers 340 steps at an effective batch of 882 completions (126 prompts $\times$ 7 rollouts) per step. Evaluation jobs round-robin $({\rm checkpoint}, {\rm task})$ pairs across the 4 GPUs of the same node, with vLLM \texttt{enforce\_eager=True}, bf16, \texttt{max\_model\_len=16384}.

\item {\bf Code of ethics}
    \item[] Question: Does the research conducted in the paper conform, in every respect, with the NeurIPS Code of Ethics \url{https://neurips.cc/public/EthicsGuidelines}?
    \item[] Answer: \answerYes{}
    \item[] Justification: The work is foundational research on training methodology for open-source language models. It does not involve human subjects, personally identifiable data, scraped web content, or sensitive applications; all training and evaluation data come from established public benchmarks released under open licenses by their original authors. We have reviewed the NeurIPS Code of Ethics and the work conforms to it.

\item {\bf Broader impacts}
    \item[] Question: Does the paper discuss both potential positive societal impacts and negative societal impacts of the work performed?
    \item[] Answer: \answerNA{}
    \item[] Justification: The contribution is a training-methodology improvement (a multi-reward RLIF objective with covariance-aware regularization) for open-source language models on math and code reasoning benchmarks. It does not introduce a new deployable system, dataset, or capability beyond what is already accessible from the underlying Qwen2.5 base models, and the evaluation domains (grade-school and competition math, code-output prediction) do not have a direct path to harmful application. We therefore do not foresee specific societal impacts beyond those already inherent in the public Qwen2.5 base models we build on.

\item {\bf Safeguards}
    \item[] Question: Does the paper describe safeguards that have been put in place for responsible release of data or models that have a high risk for misuse (e.g., pre-trained language models, image generators, or scraped datasets)?
    \item[] Answer: \answerNA{}
    \item[] Justification: The paper does not release any new high-risk asset. We fine-tune the publicly released Qwen2.5-1.5B/3B/7B-Base and Qwen3-1.7B-Base models on the public MATH-lighteval training split; any released training-recipe artifacts (configs, scripts, evaluation code) carry no additional misuse risk beyond what is already accessible from the base models and benchmark datasets.

\item {\bf Licenses for existing assets}
    \item[] Question: Are the creators or original owners of assets (e.g., code, data, models), used in the paper, properly credited and are the license and terms of use explicitly mentioned and properly respected?
    \item[] Answer: \answerYes{}
    \item[] Justification: All data, code, and methods used in this study are open-source, with proper citations and full compliance with their respective licenses and terms of use. Also all third-party assets are cited at first use and.

\item {\bf New assets}
    \item[] Question: Are new assets introduced in the paper well documented and is the documentation provided alongside the assets?
    \item[] Answer: \answerNA{}
    \item[] Justification: The paper does not release new datasets or new pre-trained models as primary assets.

\item {\bf Crowdsourcing and research with human subjects}
    \item[] Question: For crowdsourcing experiments and research with human subjects, does the paper include the full text of instructions given to participants and screenshots, if applicable, as well as details about compensation (if any)?
    \item[] Answer: \answerNA{}
    \item[] Justification: The paper does not involve crowdsourcing or research with human subjects. All training and evaluation data come from established public benchmarks (MATH, GSM8K, MMLU-Pro, AIME, LiveCodeBench, CRUXEval-O).

\item {\bf Institutional review board (IRB) approvals or equivalent for research with human subjects}
    \item[] Question: Does the paper describe potential risks incurred by study participants, whether such risks were disclosed to the subjects, and whether Institutional Review Board (IRB) approvals (or an equivalent approval/review based on the requirements of your country or institution) were obtained?
    \item[] Answer: \answerNA{}
    \item[] Justification: The paper does not involve human subjects, so IRB approval is not applicable.

\item {\bf Declaration of LLM usage}
    \item[] Question: Does the paper describe the usage of LLMs if it is an important, original, or non-standard component of the core methods in this research? Note that if the LLM is used only for writing, editing, or formatting purposes and does \emph{not} impact the core methodology, scientific rigor, or originality of the research, declaration is not required.
    %this research?
    \item[] Answer: \answerYes{}
    \item[] Justification: LLMs (Qwen2.5-1.5B-Base, Qwen2.5-3B-Base) are the \emph{subjects} of study, not used as research assistants in the development of the core methodology. The reward signals (cluster voting, self-certainty) and the KL-Cov regularizer are derived from intrinsic statistics of the policy under training and do not depend on calls to any external LLM.

\end{enumerate}

\end{document}